\documentclass[letterpaper, 10 pt, conference]{ieeeconf}  

\IEEEoverridecommandlockouts                              

\UseRawInputEncoding
\usepackage{amssymb}
\usepackage{graphicx}
\usepackage{multirow}
\usepackage[colorlinks, linkcolor=red]{hyperref}
\def\mathbi#1{\textbf{\em #1}}
\title{\LARGE \bf
Depth Completion using Geometry-Aware Embedding
}
\author{Wenchao du, Hu Chen$^{*}$, Hongyu Yang, Yi Zhang
	\thanks{This work is partly supported by the National Natural Science
	Foundation of China under grant U20A20161. W. Du, H. Chen, H. Yang and Y. Zhang are with the College of Computer Science, Sichuan University, Chengdu 610065, China. H. Chen is the corresponding author.
	{\tt\small wenchaodu.scu@gmail.com, \{huchen, yanghongyu, yzhang\}@scu.edu.cn}}%
}

\begin{document}

\maketitle
\thispagestyle{empty}
\pagestyle{empty}

\begin{abstract}

Exploiting internal spatial geometric constraints of sparse LiDARs is beneficial to depth completion, however, has been not explored well. This paper proposes an efficient method to learn geometry-aware embedding, which encodes the local and global geometric structure information from 3D points, e.g., scene layout, object's sizes and shapes, to guide dense depth estimation. Specifically, we utilize the dynamic graph representation to model generalized geometric relationship from irregular point clouds in a flexible and efficient manner. Further, we joint this embedding and corresponded RGB appearance information to infer missing depths of the scene with well structure-preserved details. The key to our method is to integrate implicit 3D geometric representation into a 2D learning architecture, which leads to a better trade-off between the performance and efficiency. Extensive experiments demonstrate that the proposed method outperforms previous works and could reconstruct fine depths with crisp boundaries in regions that are over-smoothed by them. The ablation study gives more insights into our method that could achieve significant gains with a simple design, while having better generalization capability and stability. The code is available at \href{https://github.com/Wenchao-Du/GAENet}{https://github.com/Wenchao-Du/GAENet}.
\end{abstract}
\section{INTRODUCTION}
Accurate and robust depth estimation is essential for various visual applications, such as autonomous navigation, unmanned aerial vehicles, and robotic manipulation. However, existing most of the stereo vision algorithms only suit general indoor scenes or closed areas with limited distance \cite{Szeliski2011ComputerV}, which fail for complicated outdoor scenes due to strong interference, e.g., illuminations and occlusions. The active depth sensor is an alternative solution, which provides reliable depth measurements. However, commodity-level depth sensors, e.g., LiDAR, RGBD camera, and Time-of-Flight, only produce high-resolution output with excessive sparsity, e.g., there are roughly $4\%$ valid pixels captured by Velodyne LiDAR HDL-64e on single depth map from the KITTI benchmark \cite{Uhrig2017SparsityIC}. Therefore, recovering dense depths from sparse measurements, i.e. depth completion, is critical for practical 3D vision applications.

\begin{figure}[t]
	\begin{center}
		\begin{minipage}[t]{0.97\linewidth}
			\includegraphics[width=\textwidth]{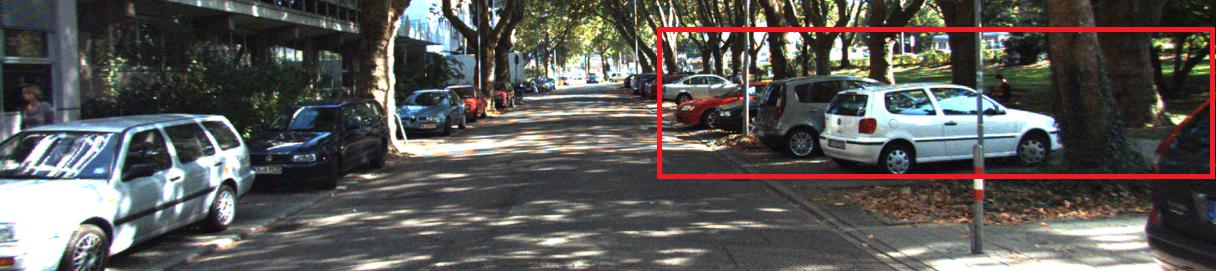}
			\centering{(a) RGB}
		\end{minipage}
		\begin{minipage}[t]{0.97\linewidth}
			\includegraphics[width=\textwidth]{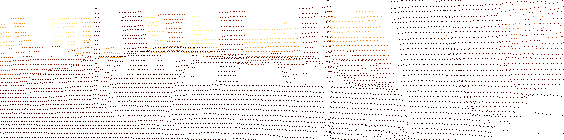}
			\centering{(b) Sparse LiDAR}
		\end{minipage}
		\begin{minipage}[t]{0.97\linewidth}
			\includegraphics[width=\textwidth]{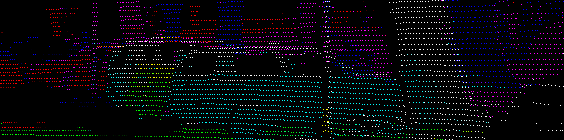}
			\centering{(c) Geometric Embedding}
		\end{minipage}
	\end{center}
	\caption{$t$-SNE visualization for the geometrical embedding. Explicit spatial topological information is lost in local sparse LiDAR data (b). Instead, the rich scene structural priors are preserved well by geometric embedding (c).}
	\label{fig1}
\end{figure}
Recent works \cite{Uhrig2017SparsityIC,Ma2018SparsetoDenseDP,Gansbeke2019SparseAN,Eldesokey2020ConfidencePT} have taken advantage of convolution neural networks (CNNs) to tackle this problem. However, excessive sparsity limits the ability of CNNs on dense grids, geometric information from sparse LiDARs is not explored well, as illustrated in Fig. \ref{fig1}. Therefore, some researches \cite{Zhang2018DeepDC,Lee2019DepthCW,Qiu2019DeepLiDARDS,Xu2019DepthCF} tried to introduce implicit normal constraints to regularize depth estimation. However, predicting accurate normal maps from complex real scenes is challenging, which limits the ability of the model. Leading works \cite{Chen2019LearningJ2,Xiang20203dDepthNetPC,Xiong2020SparsetoDenseDC} have shown that jointing 3D-2D representations extracted from point clouds and RGB images is beneficial to dense estimation. However, latent domain gap requires powerful fusion architecture, which leads to the poor generalization capabilities of model on diverse scenes. 

To tackle above problems, in this paper, we focus on two key factors, i.e. better geometric representation and efficient fusion architecture. Our goal is to exploit internal geometric structure relationship to regularize depth estimation in a simple but effective manner. As shown in Fig. \ref{fig1}, we observe that the geometric properties of objects, e.g., size and shape, are still preserved in 3D space, which provide reliable priors for structural scene perception while having better robustness for occlusions and illuminations. To this end, we aim to explore generalized geometric representation from 3D point clouds, which encodes the local and global structure information (reference as \textit{geometry-aware embedding}), and further fuses the 2D appearance features from RGB images to estimate dense depth, where geometric embedding could be viewed as implicit auxiliary constraints to regularize the depth prediction. Therefore, the proposed method contains two modules, i.e. learning geometry-aware embedding and geometry-guided propagation, they are integrated into a unified reasoning framework. In this way, the visual and geometrical information can be exploited well to regularize depth reconstruction, leading the final prediction to be both structural and geometric consistency.

Our \textbf{contributions} are summarized as follows: (1) Propose a novel approach that learns geometric-aware embedding from sparse LiDARs via task-driven without auxiliary supervision, to regularize depth inference effectively; (2) Design a simple yet efficient joint framework for depth completion that integrates 3D-2D information better and is end-to-end trainable; (3) Achieve finer depth estimation in indoor and outdoor scenes while having better generalization capability and stability.

\section{Related Work}

\textbf{RGB-D Depth Completion}. Early works viewed depth completion as a typical inverse problem, some classical image processing methods are applied to it, e.g., compressed sensing \cite{Hawe2011DenseDM} and wave analysis \cite{Liu2015DepthRF}. Benefited from CNNs' evolution on dense grids, some methods have tried to extend CNNs' operators to sparse LiDAR maps \cite{Uhrig2017SparsityIC,Huang2020HMSNetHM}.

Inspired by depth estimation from the single RGB image, Ma and Karaman \cite{Ma2018SparsetoDenseDP} directly fused the RGB-LiDAR to predict dense depth by an encoder-decoder network and achieved significant improvements. Cheng et al. \cite{Cheng2018DepthEV} introduced a recurrent convolution into spatial propagation network (reference as CSPN) to estimate the affinity matrix as a post-processing module, which speeds the SPN \cite{Liu2017LearningAV} and improves the performance. Park et al. \cite{Park2020NonLocalSP} further applied the non-local convolutions to CSPN, enhancing the ability on learning robust affinity matrix. Along this direction, recent works \cite{Cheng2020CSPNLC,Hu2021PENetTP} focus on improving performance of CSPN by careful designing. But they all require powerful backbone to learn affinity matrix and sufficient iterations for finer refinement, which lead to the slow inference speed. In addition, some researches also explored better RGB-D fusion architecture for dense estimation \cite{Shivakumar2019DFuseNetDF,Gansbeke2019SparseAN,Tang2021LearningGC}.

\textbf{Geometry-Aware Depth Completion}. Recently, exploiting implicit geometric constraints to regularize dense estimation has drawn more attention. Zhang et al. \cite{Zhang2018DeepDC} firstly utilized normal constraints to guide depth completion in indoor scenes. Further, Qiu et al. \cite{Qiu2019DeepLiDARDS} extended normal estimation as an intermediate representation for outdoor depth prediction. Similarly, Xu et al. \cite{Xu2019DepthCF} constructed plane-origin distance constraints based on predicted normal map to refine final depths. However, a key point is ignored by them that estimating accurate normal from outdoor scenes is challenging due to uncertain interference. In addition, some works \cite{Imran2021DepthCW,Lee2021DepthCU} exploited plane constraints for fine depth completion. Furthermore, view synthesis based self-supervised methods \cite{Ma2019SelfSupervisedSS,Wong2020UnsupervisedDC} have been also explored. 

Considering that the latent topology information is lost in 2D space, fusing 2D-3D features from RGB-Point clouds for dense estimation has shown great advantages in recent work. Chen et al. \cite{Chen2019LearningJ2} first constructed 3D-2D fusion blocks to aid for depth estimation, significant performance gains are achieved by increasing the width and depth of the model. Xiang et al. \cite{Xiang20203dDepthNetPC} introduced an extra point clouds completion network into general encoder-decoder architecture. Although these methods have exploited 3D-2D representation, implicit geometric constraints are not explored well. In addition, large-scale point clouds require heavy computation budgets during the training and testing, so that they could not achieve a better trade-off between the performance and efficiency.

\textbf{Graph Representation Learning}. Following the breakthrough progress of CNN in dense grids, some researches \cite{Wang2018DeepPC,Wu2019PointConvDC} have applied it to irregular data, e.g., point clouds. Previous works transform the point cloud onto a grid, e.g., view-based \cite{Su2015MultiviewCN} and volumetric representations \cite{Klokov2017EscapeFC}. PointNet series \cite{Qi2017PointNetDL,Qi2017PointNetDH} exploited the point-wise MLP to learn spatial features from 3D points. Recently, the graph has been viewed as a type of non-Euclidean structure and used to represent unordered data. Graph neural networks directly define convolutions in the spectral and spatial domains, operating on groups of spatially close neighbors to exploit the geometric relationship \cite{Te2018RGCNNRG}, have made significant progress on point cloud classification and segmentation \cite{Wang2019DynamicGC,Zhang2019LinkedDG}. For dense estimation, Xiong et al. \cite{Xiong2020SparsetoDenseDC} explored graph construction on 3D points to guide depth completion. Further, Zhao et al. \cite{Zhao2021AdaptiveCM} directly constructed an encoder-decoder architecture based on graph propagation to explore better multi-modality feature fusion for depth completion. Although graph-based methods generally have fewer parameters, large-scale graph construction and messages aggregation require heavy computations, which limits general applications.

\begin{figure*}
	\begin{center}
		\includegraphics[width=1.0\linewidth]{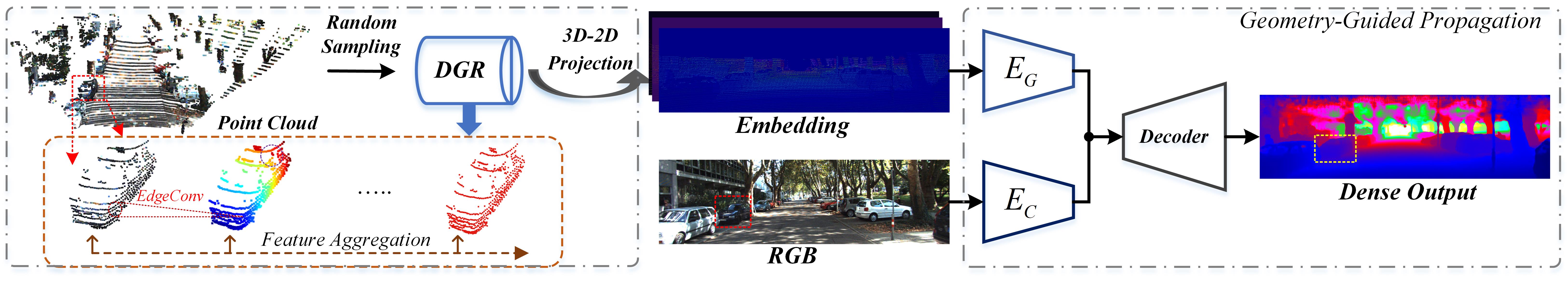}
	\end{center}
	\caption{Overview of the method. Our method first extracts the geometric embedding from sparse point clouds by a dynamic graph representation (DGR) module, and then uses this embedding to guide dense reconstruction. The whole framework is injected into a general encoder-decoder architecture.}
	\label{fig2}
\end{figure*}
\section{Method}

This section describes details of the proposed method, as shown in Fig. \ref{fig2}, which contains two modules, i.e. geometry representation learning and geometry-guided propagation.

\subsection{Learning Geometric-Aware Embedding} 

\textbf{Dynamic Graph Representation (DGR)}. To exploit the spatial geometric relationship in point cloud space, we introduce the edge convolution (EdgeConv) \cite{Wang2019DynamicGC} as the basic component of DGR module to capture local geometrical information while maintaining permutation invariance.

Given a point cloud with $n$ points is denoted by $\mathbf{X}=\{\mathbf{x_1},\dots,\mathbf{x_n}\}\subseteq \mathbb{R}^F$, where $F=3$ means 3D coordinates are contained in each point that $\mathbf{x}_i=(x_i,y_i,z_i)$. A directed graph $\mathcal{G}=(\mathcal{V}, \mathcal{E})$ is used to model local spatial relationships, where $\mathcal{V}={1,\dots,n}$ and $\mathcal{E} \subseteq \mathcal{V} \times \mathcal{V}$ denote the vertices and edges, respectively. $\mathcal{G}$ is constructed by the $k$-nearest neighbor ($k$-NN) algorithm in $\mathbf{X}\subseteq \mathbb{R}^{3}$. The single EdgeConv is defined as
$$
\mathbf{x}'_i=\max\limits_{j:(i,j)\subseteq\mathcal{E}}h_{\Theta}(\mathbf{x}_i,\mathbf{x}_j)\eqno{(1)}
$$
$$
h_{\Theta}(\mathbf{x}_i,\mathbf{x}_j)=\hat{h}_{\Theta}(\mathbf{x}_i,\mathbf{x}_j-\mathbf{x}_i)\eqno{(2)}
$$
where $\mathbf{x}_i$ is regarded as central point and $j:(i,j)\subseteq\mathcal{E}$ defines a patch around it. $h_{\Theta}: \mathbb{R}^{F}\times \mathbb{R}^{F}\rightarrow \mathbb{R}^{F'}$ is a nonlinear edge function with learnable parameters $\Theta$. It produces the edge features associating with all the edges emanated from the vertex $\mathbf{x}_i$. $\max\{\cdot\}$ defines an aggregation operation on the edge features, which is channel-wise symmetric so that $\mathbf{x}'_i$ is invariant to permutation of the input $\mathbf{x}_j$. Given an $F$-dimensional point cloud with $n$ points, EdgeConv produces an $F'$-dimensional point cloud with the same number of points. Further, the edge function is reformulated as (2), which explicitly combines global shape structure, captured by the $\mathbf{x}_i$ with local neighborhood information, captured by $\mathbf{x}_j - \mathbf{x}_i$ while preserving translation invariant property.

At each layer, a different graph $\mathcal{G}^{(l)}=(\mathcal{V}^{(l)},\mathcal{E}^{(l)})$ is constructed by computing a pairwise distance matrix in feature space and the closest $k$ neighbors are selected for each point. Such a dynamic graph construction strategy leads to our DGR module, which explores the proximity in feature space instead of fixed input, results in the more generalized and stable geometric structure representation. In practice, the single EdgeConv is implemented by one shared linear convolution layer and followed by a batch normalization and a ReLU non-linearity.

\textbf{Local and Global Feature Aggregation}. DGR could capture local geometric relationships in each layer. However, with dynamic graph updating, the receptive field of single point feature would be as large as the diameter of point clouds, which leads that the point feature from last layer tends to capture global structural relationship but loses the local geometrical information. This is unsuitable for robust geometric representation. Thus, a reduction function $\mathcal{R}(\cdot)$ is designed to encode the local and global geometrical information for each point $\textbf{x}_i$, which is defined as follows:
$$
{\mathbf{F}_{i}} = \mathcal{R}\left(f_{i}^{1} \odot f_{i}^{2}\odot \cdots\odot f_{i}^{l} \cdots \odot f_{i}^{L}, \mathbi{W}\right)\eqno{(3)}
$$
where $f_{i}^{l}$ denotes the point feature from the $l$-th layer of the center point $\mathbf{x}_i$, the output from each layer of $\mathbf{x}_i$ forms a feature vector $[f_{i}^{1}, f_{i}^{2}, \cdots, f_{i}^{l}, \cdots, f_{i}^{L}]$, which contains the local and global geometric information. For initial input, i.e. $f_{i}^{0}=\mathbf{x}_i$, it means $f_{i}^{0}$ only contains explicit 3D coordinate information. We use a shared linear convolution layer as the aggregation function to encode multi-scale geometric feature vectors with learnable weights $\mathbi{W}$, $\odot$ defines a channel-wise concatenation operation. Here we stack four EdgeConv layers, i.e. $\mathcal{L}=4$, to extract multi-scaled point-wise features with multiple dimensions $\left[64, 64, 128, 256\right]$. Furthermore, we use a Max-Pooling operator to acquire the final feature vector ${\mathbf{F}_{i}}$, $Size(\textbf{F}_i)=16$. In addition, considering the explicit depth clues are lost in $\textbf{F}_i$, it is unreliable that relying on the implicit geometric constraints only to guide depth estimation. Therefore, we further concatenate the depth clue $z_{i}$ with $\mathbf{F}_{i}$ to form the final representation $\hat{\mathbf{F}}_{i}$, referred as \textit{geometry-aware embedding}, where $\hat{\mathbf{F}}_{i}=\mathbf{F}_{i}\odot{z_i}$, $Size(\hat{\mathbf{F}}_{i})=17$.

During the training and testing, DGR first transforms the sparse depth map into point clouds using known camera intrinsic parameters, and then extracts the geometric embedding of each point in 3D space, so that we get the 3D points of shape $N\times 3$ along with point features of shape $N\times C$, where $C=17$. After this step, we back-project the $N\times 3$ 3D points to an empty 2D feature map and assign the $N\times C$ point features to corresponding projected pixels. In this way, we obtain a sparse 2D feature map as the output of the DGR module, which has the same shape as the sparse depth map but with $C$ channels. Therefore, the single valid pixel in 2D sparse feature map not only carries the explicit depth information, but it also contains the implicit geometric clues from local and global spatial structures.
\begin{figure*}[thpb]
	\centering
	\begin{minipage}[t]{0.325\linewidth}
		\centerline{\includegraphics[width=\textwidth]{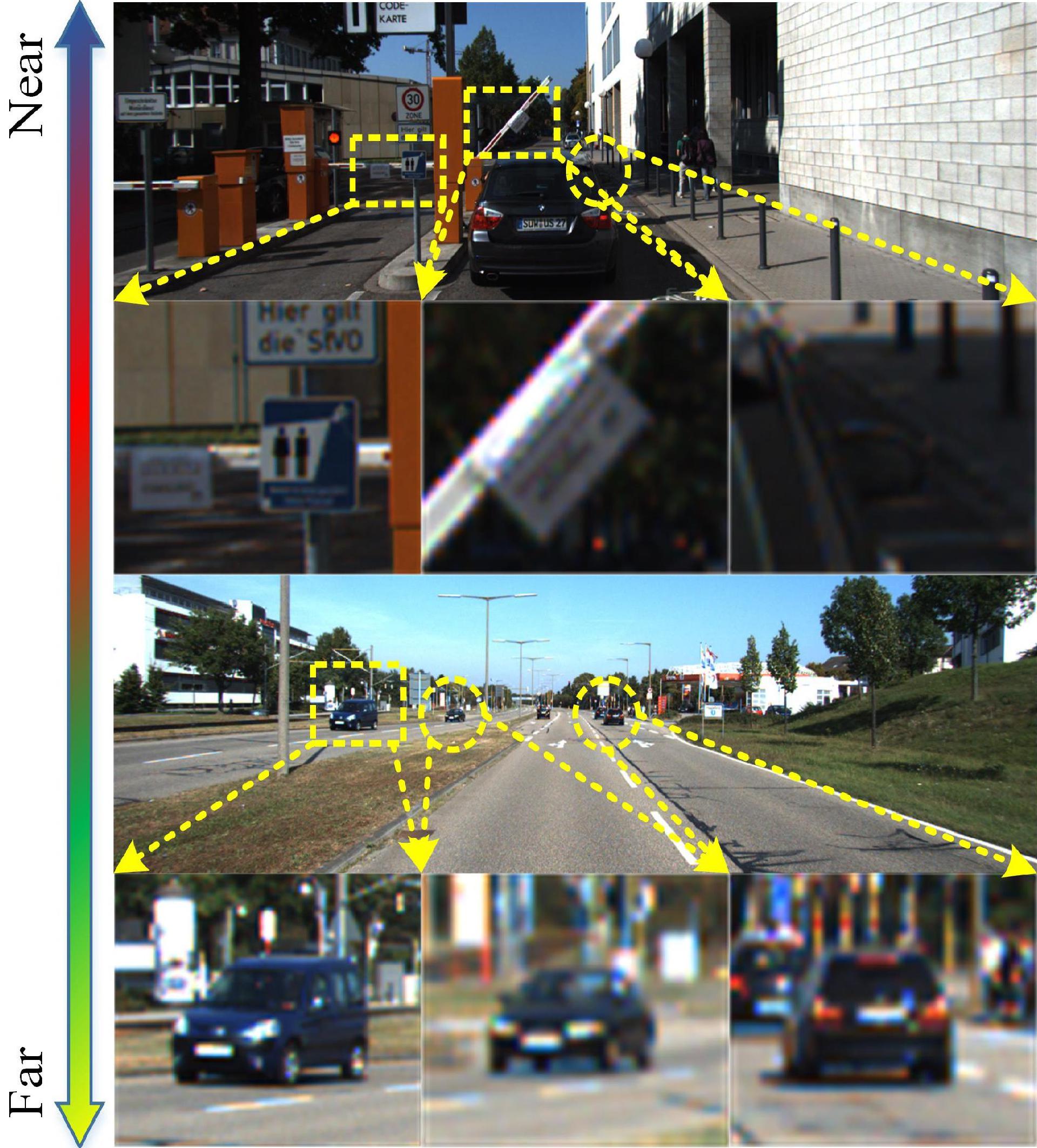}}
		\centerline{RGB}
	\end{minipage}
	\begin{minipage}[t]{0.325\linewidth}
		\centerline{\includegraphics[width=\textwidth]{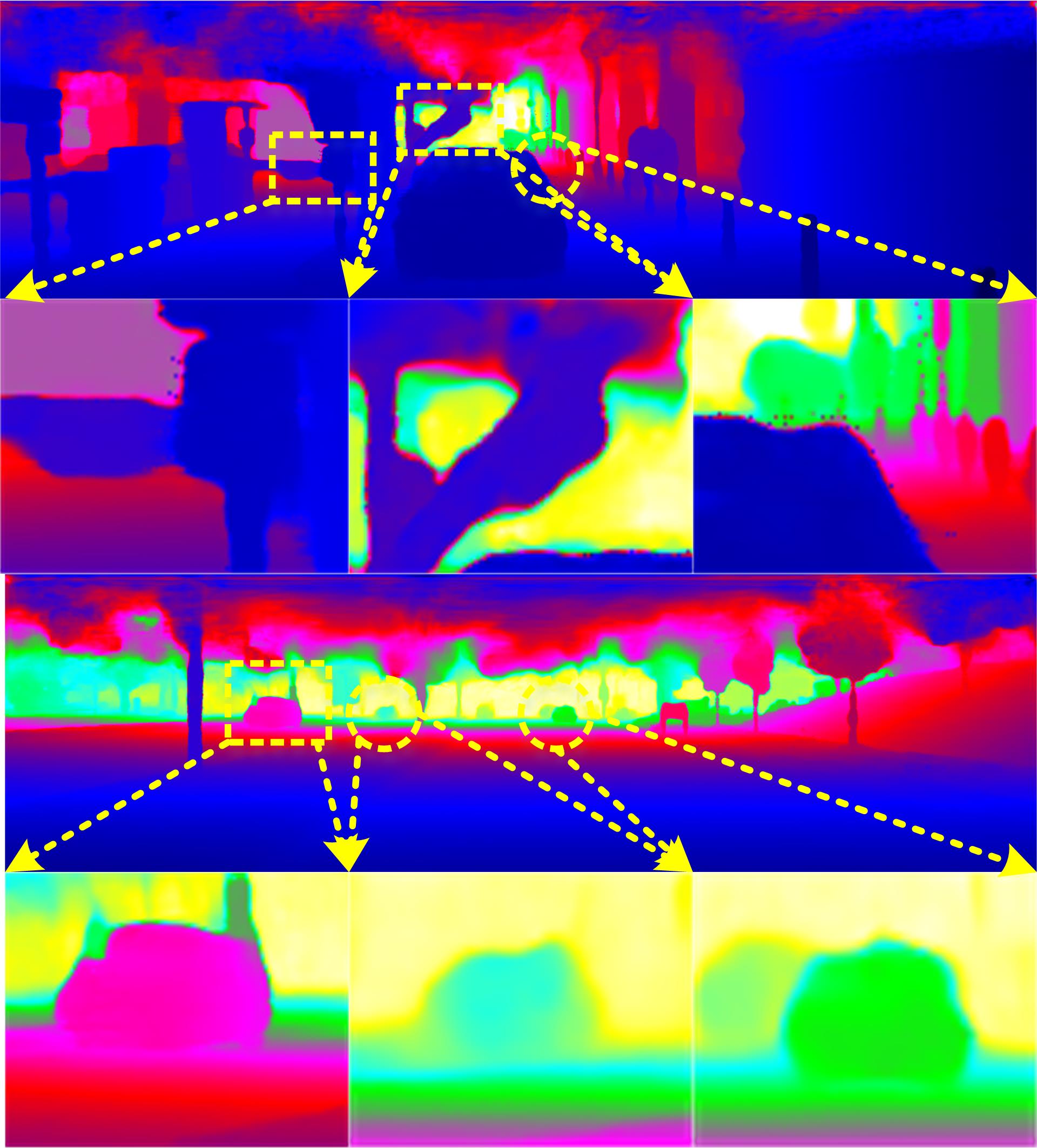}}
		\centerline{Uber-FuseNet \cite{Chen2019LearningJ2}}
	\end{minipage}
	\begin{minipage}[t]{0.325\linewidth}
		\centerline{\includegraphics[width=\textwidth]{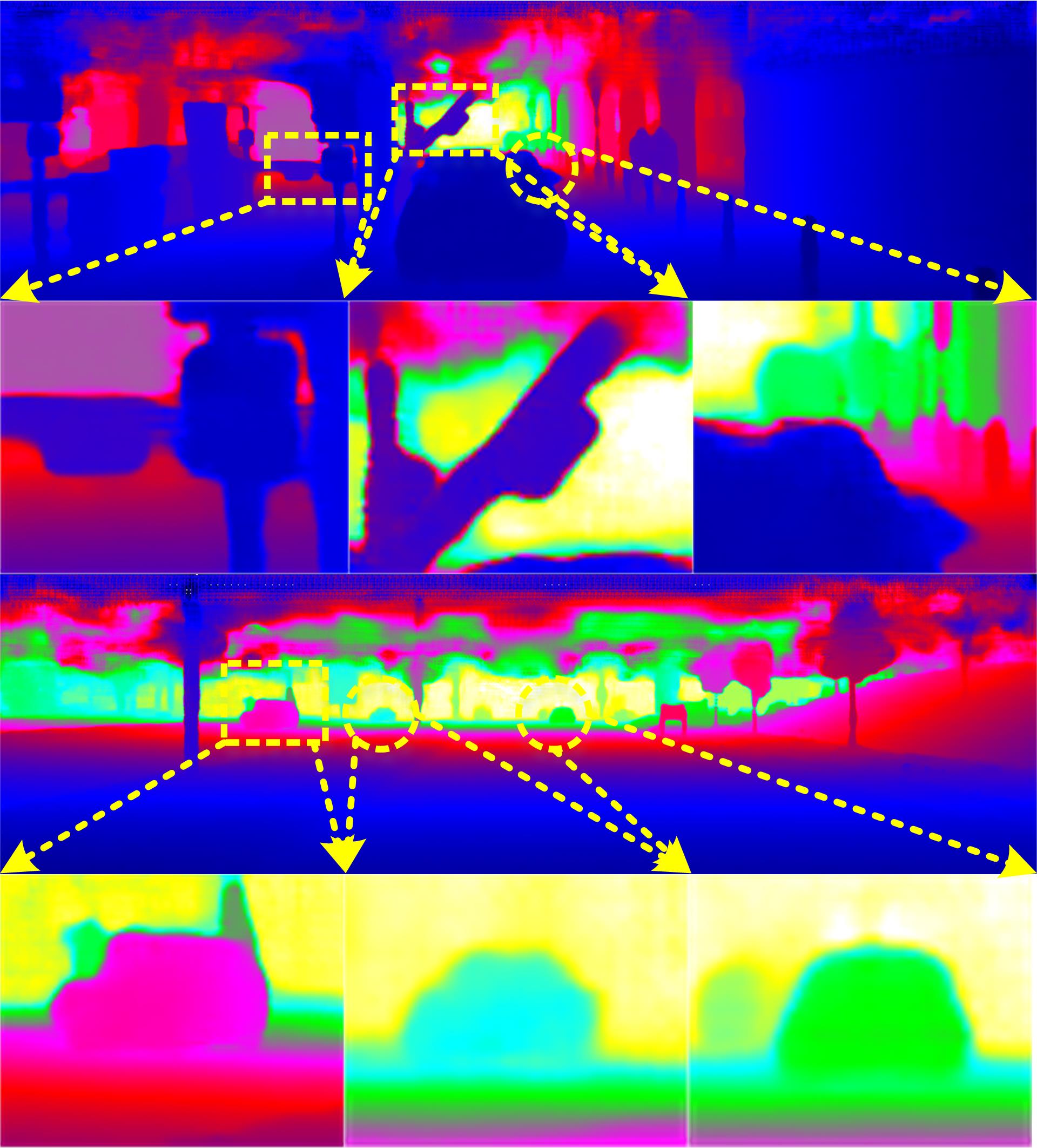}}
		\centerline{Ours}
	\end{minipage}
	\caption{Visualized results from the KITTI test set. Uber-FuseNet generates over-smoothed results in local areas containing rich structures. In contrast, our method retains the finer objects' properties, e.g., edges and shapes, and presents more robust estimation for objects in wide ranges.}
	\label{fig3}
\end{figure*}
\subsection{Geometry-Guided Propagation}

Although monocular depth prediction \cite{Zhao2020MonocularDE} is a typical ill-posed problem due to latent uncertainty, it implies the rich visual appearance from single RGB image could provide extra information for depth perception. Hence, we propose a propagation framework to leverage the learned embedding as guidance, which utilizes visual and geometric information to infer missing depths.

The bi-branch encoding strategy \cite{Qiu2019DeepLiDARDS,Wong2020UnsupervisedDC,Jaritz2018SparseAD} is used in our framework, which contains two independent encoders, referred as $\mathbi{E}_{C}$ and $\mathbi{E}_{G}$. $\mathbi{E}_{C}$ aims to extract multi-scaled visual features from the color image $I_{C}$. For $\mathbi{E}_{G}$, we don't only expect to acquire dense representation from the sparse geometric embedding $I_{G}$, but also attempt to explore the multi-scale geometric constraints from 2D sparse depths.

In practice, our encoders down-sample inputs to $1/8$ scale so that we get the multi-scale feature maps with $[1/2, 1/4, 1/8]$, they contain rich visual and geometric information, and then are used for dense reconstruction.

Furthermore, to exploit the visual and geometric features from different scales well, we define a simple but effective upsampling operation as
$$
\mathbf{p}^{l-1}={g_U}_{l}((\mathbf{p}^{l}\oplus{f}_{C}^{l})\odot {f}_{G}^{l}, \mathbi{W}_{l}^{T})\eqno{(4)}
$$
where ${g_U}_{l}(\cdot)$ is an upsampling function for scale $l$, $\mathbf{p}^{l}$ is the upsampled feature map from the $l$-{th} scale, which is first fused with feature map ${f}_{C}^{l}$ extracted from $I_C$ with a point-wise summation operation $\oplus$, where ${f}_{C}^{l}$ provides dense visual appearance information. Further, we concatenate ${f}_{G}^{l}$, which provides implicit geometric constraints containing depth and spatial structure information to guide upsampling. This is an efficient way to enforce the visual and geometric constraints to regularize the depth estimation. $\mathbi{W}_{l}^{T}$ is the learnable parameter. For the initial upsampled feature map, i.e. $L=3$, which is written as $\mathbf{p}^{L-1}={g_U}_{L}({f}_{C}^{L} \odot {f}_{G}^{L}, \mathbi{W}_{L}^{T})$. The final output could be denoted as
$\mathbf{p}^{0}={g_U}_{1}((\mathbf{p}^{1}\oplus f_{C}^{1})\odot f_{G}^{1}, \mathbi{W}_{1}^{T})
$. In practice, we implement the $g_U(\cdot)$ with a general transposed convolution. 

The whole framework is shown in Fig. \ref{fig2}. To validate the efficiency of our method while balancing computation costs further, we use the ERFNet \cite{romera2018erfnet} as backbone of our framework, which is more lightweight than MobileNetv2 \cite{Sandler2018MobileNetV2IR}. We modify the last layer of it to regress dense depth instead of classification and an extra encoding branch is added as ${E}_{G}$. For convenience, ${E}_{G}$ uses a same network structure as ${E}_{C}$ but with different input channels.

\subsection{Loss Function}
Considering that our approach is integrated into a unified framework, it could be trained with end-to-end, and does not require any special training strategies and loss functions like previous methods \cite{Ma2018SparsetoDenseDP,Chen2019LearningJ2}. The mean squared error (MSE) loss function is used only, which is denoted by
$$
\mathcal{L}_{MSE}=\sum_{v\in\mathcal{V}}\parallel \mathbf{P}_{gt}^{v}-\mathbf{P}^{v}\parallel^{2}\eqno{(5)}
$$
where $\mathcal{V}$ represents the set of valid pixels in the LiDAR map. $\mathbf{P}_{gt}^{v}$ and $\mathbf{P}^{v}$ denote the ground truth and predicted depth at the pixel $v$ respectively.
\begin{table}[t]
	\begin{center}
		\setlength{\tabcolsep}{1.7mm}
		\scalebox{0.95}{
			\begin{tabular}{|l|cccc|c|}
				\hline
				Method & \textbf{RMSE}$\downarrow$ & MAE$\downarrow$   & iRMSE$\downarrow$  & iMAE$\downarrow$  &\textit{Params} \\
				\hline\hline
				SparseConvs	\cite{Uhrig2017SparsityIC}	& $1601.3$ & $481.27$ & $4.94$ & $1.78$ & $0.026$M \\
				CSPN \cite{Cheng2018DepthEV} 			& $1019.6$  & $279.46$ & $2.93$ & $1.15$ &  $172.8$M\\
				Sparse2Dense \cite{Ma2019SelfSupervisedSS}  & $814.73$  & $249.95$ & $2.80$ & $1.21$ &  $26.10$M\\
				DeepLiDAR \cite{Qiu2019DeepLiDARDS}  & $\textbf{758.38}$  & $226.50$ & $2.56$ & $1.15$ &  $144.0$M\\
				DepthNormal \cite{Xu2019DepthCF} & $777.05$  & $235.17$ & $2.42$ & $1.13$  & --\\
				Uber-FuseNet \cite{Chen2019LearningJ2} & $\textbf{752.88}$  & $221.19$ & $2.34$ & $1.14$ & $\textbf{1.898}$M \\
				3dDepthNet \cite{Xiang20203dDepthNetPC} & $798.44$ & $226.27$ & $2.36$ & $1.02$& --\\
				TWISE \cite{Imran2021DepthCW} & $829.60$  & $196.88$ & $2.09$ & $0.84$&	$1.450$M\\
				\textbf{Ours}  		& \textbf{773.90}  & \textbf{231.29} & \textbf{2.29} & \textbf{1.08}& \textbf{4.189}M \\
				\hline
			\end{tabular}
		}
	\end{center}
	\caption{Quantitative results of state-of-the-art methods on KITTI depth completion test set.}
	\label{tab1}
\end{table}
\section{Experiments}

We implement the proposed method with Pytorch on the single Nvidia TITAN GPU. Our models are trained on KITTI \cite{Uhrig2017SparsityIC} and NYUv2 \cite{Silberman2012IndoorSA} data sets. Adam \cite{Kingma2015AdamAM} is used as the optimizer. Considering more challenges in large-scale outdoor scenes, the batch size is set 4 and initial learning rate is set $2e-3$ for KITTI dataset with 50 training epochs. As for NYUv2, we increase the batch size to 16 with initial learning rate $1e-2$, and train the network with 30 epochs only, the learning rate drops by half every 10 epoch iterations. Besides, extra horizontal flipping is used to augment training set. Specially, we set $k=9$ to construct the dynamic graph in our DGR module to balance the performance and computational costs.

\subsection{Dataset and Metrics}

\textbf{KITTI Dataset}. The KITTI depth completion benchmark \cite{Uhrig2017SparsityIC}, a large self-driving real-world dataset with street views from a driving vehicle, contains 86,896 frames for training, 1000 frames for selected validation, and 1000 frames for testing. Considering image size in KITTI validation and test sets is $1216\times352$, we randomly crop
the training image to the same size. For DGR module, to speed training while improving the generalization capacity of the model, we randomly sample 8,000 3D points to learn geometrical embedding during training.

\textbf{NYUv2 Dataset}. NYUv2 Depth Dataset \cite{Silberman2012IndoorSA} consists of paired RGB-Depth images captured by Microsoft Kinect in 464 indoor scenes. Following the same training setting of previous methods \cite{Ma2018SparsetoDenseDP,Zhang2018DeepDC}, we sample about $43k$ paired RGB-Depth images from training set, and evaluate on the $654$ official labeled test set. Moreover, preprocessing is performed with the official toolbox. The original frames of size $640\times 480$ are half down-sampled with bilinear interpolation, and then center-cropped to $304\times 224$.

\textbf{Evaluation Metrics}. Four typical metrics are adopted in the KITTI evaluation, which are Root Mean Square Error (RMSE[mm]), Mean Absolute Error (MAE[mm]), inverse RMSE(iRMSE[1/km]), and inverse MAE(iMAE[1/km]) respectively. For the NYUv2, we use the same evaluation metrics with previous methods \cite{Ma2018SparsetoDenseDP,Qiu2019DeepLiDARDS}, which are RMSE(m), mean absolute relative error (REL) and $\delta_i$ which means the percentage of predicted pixels where the relative error is less a threshold $i$. Specifically, $i$ is chosen as $1.25$, $1.25^2$ and $1.25^3$ separately. Here the RMSE is used as the primary metric in all the quantitative evaluations.
\begin{table}[t]
	\begin{center}
		\scalebox{0.8}{
			\begin{tabular}{|c|l|cc|ccc|} 
				\hline
				Samples & Method & RMSE$\downarrow$ & REL$\downarrow$ & $\delta_{1.25}$$\uparrow$ & $\delta_{1.25^2}$$\uparrow$ & $\delta_{1.25^3}$$\uparrow$ \\
				\hline\hline
				\multirow{8}{*}{500} & TGV \cite{Ferstl2013ImageGD} & $0.635$ & $0.123$ & $81.9$ & $93.0$ & $96.8$ \\
				& Zhang et al. \cite{Zhang2018DeepDC} & $0.228$ & $0.042$ & $97.1$ & $99.3$ & $99.7$ \\
				& Sparse2Dense \cite{Ma2018SparsetoDenseDP} & $0.204$ & $0.043$ & $97.8$ & $99.6$ & $99.9$ \\
				& CSPN \cite{Cheng2018DepthEV}  & $0.117$ & $0.016$ & $99.2$ & $99.9$ & $100$	\\
				& DeepLiDAR \cite{Qiu2019DeepLiDARDS} & $0.115$ & $0.022$ & $99.3$ & $99.9$ & $100$ \\
				& DepthNormal \cite{Xu2019DepthCF} & $0.119$ & $0.021$ & $\mathbf{99.4}$ & $99.9$ & $100$\\
				& \textbf{Ours(w/o DGR)} & $\mathbf{0.124}$ & $\mathbf{0.021}$ & $99.1$& $\mathbf{99.9}$& $\mathbf{100}$\\
				& \textbf{Ours(w/ DGR)} & $\mathbf{0.114}$ & $\mathbf{0.018}$ & $99.3$& $\mathbf{99.9}$& $\mathbf{100}$\\
				\hline
				\multirow{4}{*}{200} & Sparse2Dense \cite{Ma2018SparsetoDenseDP} & $0.230$ & $0.044$ & $97.1$ & $99.4$ & $99.8$\\
				& NConv-CNN \cite{Eldesokey2020ConfidencePT} & $0.173$ & $0.027$ & $98.2$ & $99.6$ & $99.9$\\
				& \textbf{Ours(w/o DGR)} & $\mathbf{0.159}$ & $\mathbf{0.027}$ & $\mathbf{98.6}$ & $\mathbf{99.7}$& $\mathbf{99.9}$\\
				& \textbf{Ours(w/ DGR)} & $\mathbf{0.157}$ & $\mathbf{0.026}$ & $\mathbf{98.7}$ & $\mathbf{99.7}$& $\mathbf{99.9}$\\
				\hline
			\end{tabular}
		}
	\end{center}
	\caption{
		The quantitative comparisons with state-of-the-arts on NYUv2 test set.
	}
	\label{tab2}
\end{table}
\begin{figure}[t]
	\begin{center}
		\begin{minipage}[t]{0.18\linewidth}
			\centerline{\includegraphics[width=\textwidth]{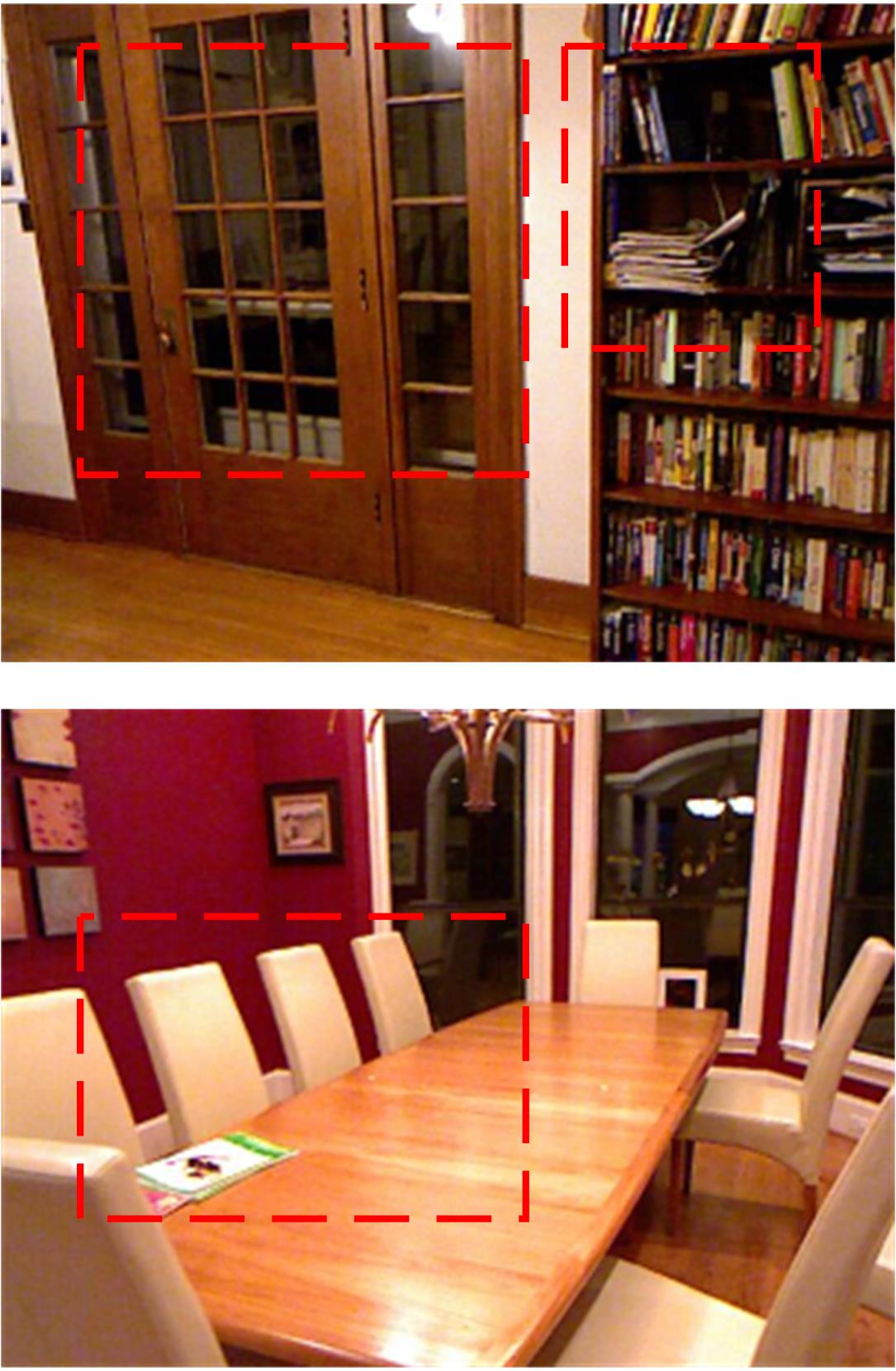}}
			\centerline{RGB}
			\centerline{(Samples)}
		\end{minipage}
		\begin{minipage}[t]{0.18\linewidth}
			\centerline{\includegraphics[width=\textwidth]{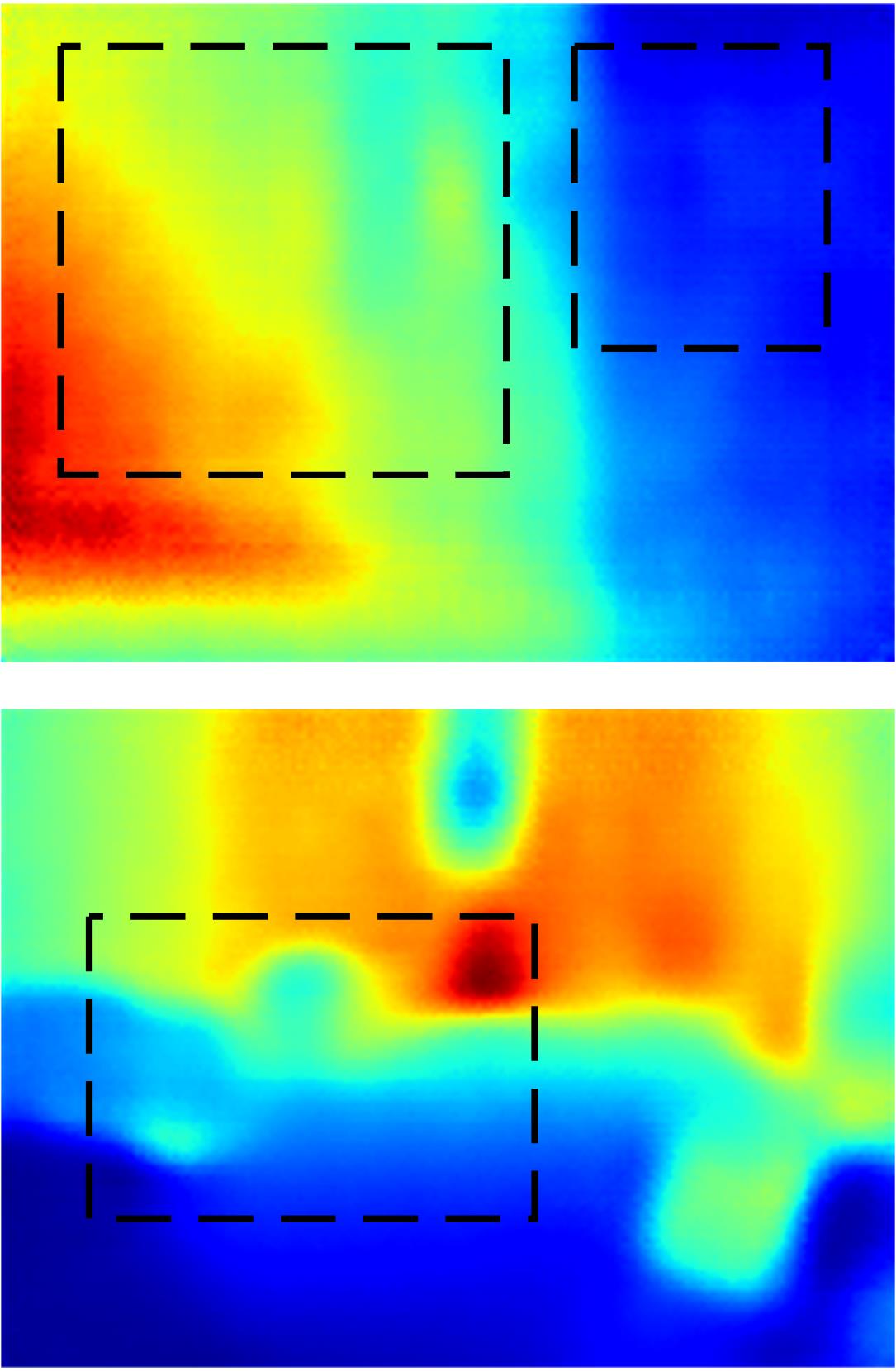}}
			\centerline{Sparse2Dense}
			\centerline{(200)}
		\end{minipage}
		\begin{minipage}[t]{0.18\linewidth}
			\centerline{\includegraphics[width=\textwidth]{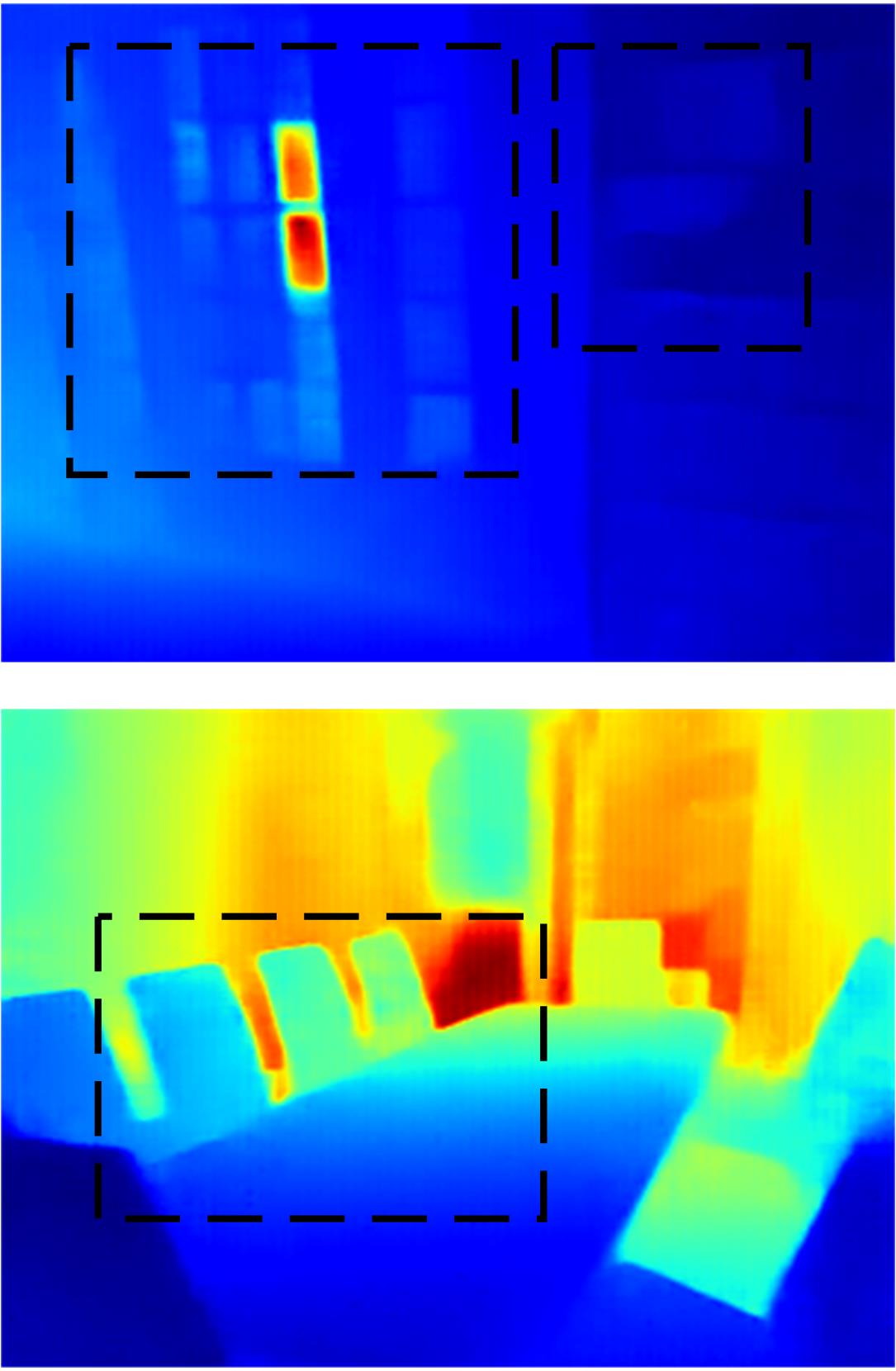}}
			\centerline{Ours}
			\centerline{(200)}
		\end{minipage}
		\begin{minipage}[t]{0.18\linewidth}
			\centerline{\includegraphics[width=\textwidth]{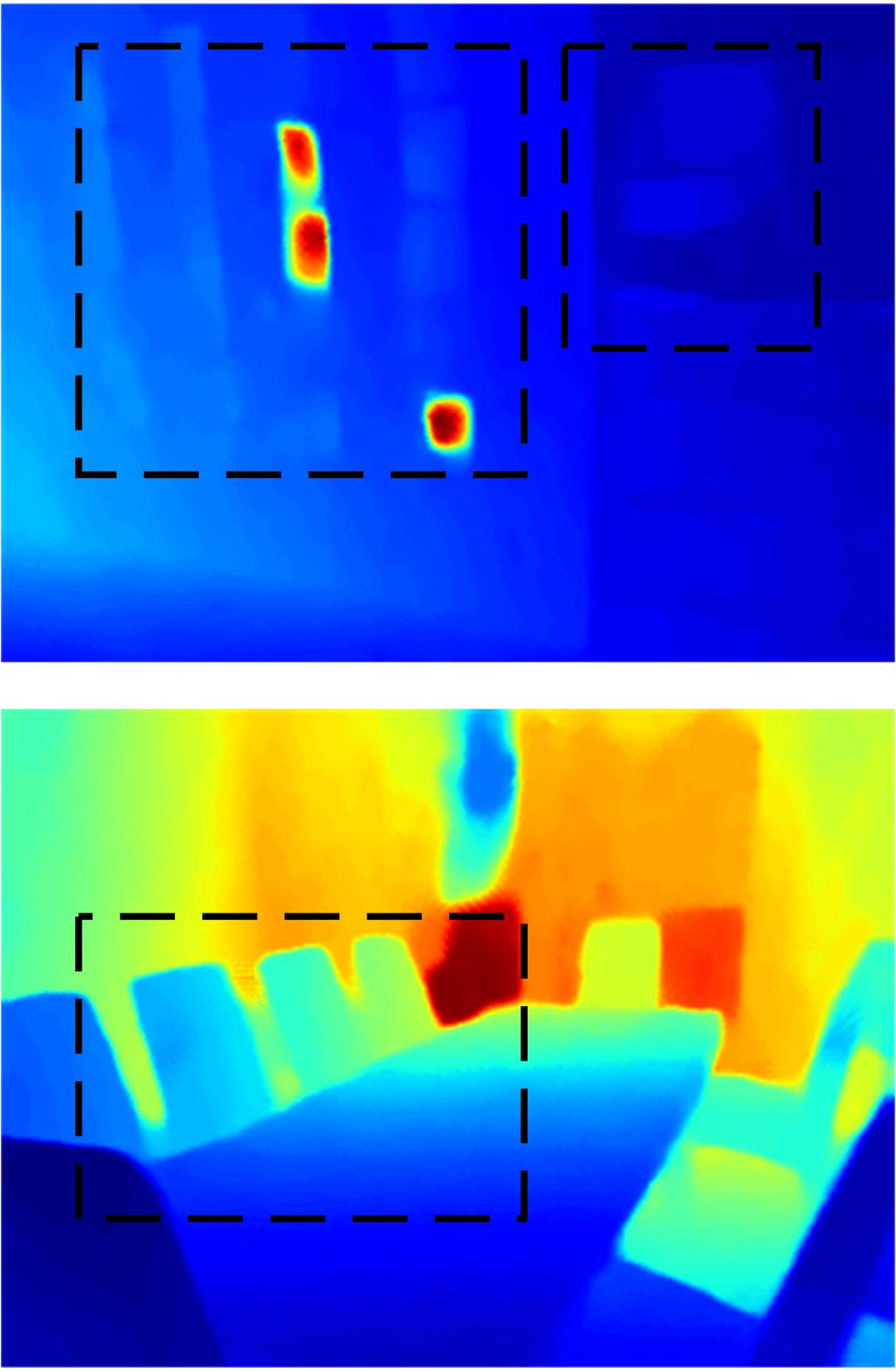}}
			\centerline{CSPN}
			\centerline{(500)}
		\end{minipage}
		\begin{minipage}[t]{0.18\linewidth}
			\centerline{\includegraphics[width=\textwidth]{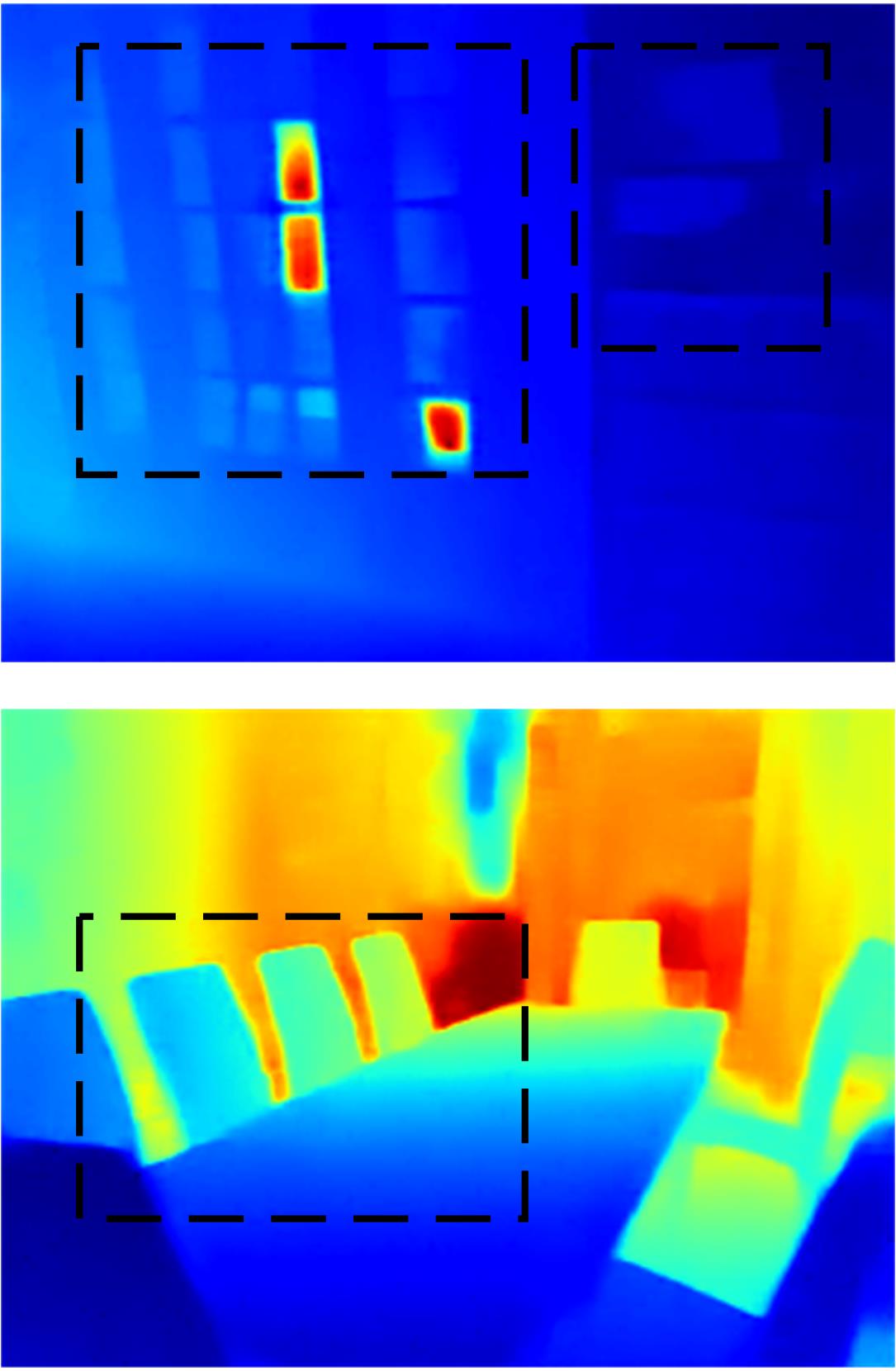}}
			\centerline{Ours}
			\centerline{(500)}
		\end{minipage}
	\end{center}
	\caption{Qualitative comparison with state-of-the-arts on NYUv2 test set. The proposed method presents more reliable ability on structural scene perception and achieves finer depth reconstruction with fewer training parameters against other methods.}
	\label{fig4}
\end{figure}
\begin{table}[t]
	\begin{center}
		\setlength{\tabcolsep}{1.7mm}
		\scalebox{0.95}{
			\begin{tabular}{|l|lccc|l|} 
				\hline
				Variants & \textbf{RMSE}$\downarrow$& MAE$\downarrow$ & iRMSE $\downarrow$& iMAE$\downarrow$ & \#Params(\textbf{M})\\
				\hline\hline
				Baseline & $844.169$& $243.247$ & $2.669$ & $1.065$ & 2.094\\
				w/ BBE & $834.464$& $244.717$ & $2.715$ & $1.239$ & 4.063$(-)$\\
				w/ SG & $829.766$& $254.76$ & $2.573$ & $1.189$ & 4.196$(+0.133)$\\
				w/o FA & $826.165$& $248.569$ & $2.464$ & $1.131$ & 4.156$(+0.093)$\\
				Full+FPS & $823.778$& $237.767$ & $2.513$ & $1.088$ & 4.189$(+0.126)$\\
				Full+RS & $\textbf{813.831}$& $245.081$ & $2.660$ & $1.225$ & 4.189$(+0.126)$ \\
				\hline
			\end{tabular}
		}
	\end{center}
	\caption{
		The quantitative results of different ablation variants on KITTI selected validation set.
	}
	\label{tab3}
\end{table}
\subsection{Comparison with State-of-the-arts}

\textbf{Results on KITTI}. We first evaluate our method on the test set of the KITTI by submitting results to KITTI evaluation server, Table \ref{tab1} lists the detailed results. Compared to general methods, e.g., CSPN and Sparse2Dense, our method achieves superior improvements on metrics. For normal constraints based methods \cite{Qiu2019DeepLiDARDS,Xu2019DepthCF}, our method achieves the better trade-off between the performance and efficiency. Compared with 3dDepthNet \cite{Xiang20203dDepthNetPC}, our method only injects geometrical features into a general framework and achieves better metrics. Uber-FuseNet \cite{Chen2019LearningJ2} has the lower RMSE and fewer parameters against ours, but it requires heavy computational costs due to repeatedly graph propagation on full-scale point clouds. Instead, our method only extracts 3D representation once to refine depth estimation with a coarse to fine manner, which is more efficient. Further, our DGR module explores the structure constraints from 3D space, it couldn't be exploited well due to less structural priors in certain outdoor scenes. However, our method still outperforms other methods, e.g., DepthNormal \cite{Xu2019DepthCF} and TWISE \cite{Imran2021DepthCW}.

Visualized results are shown in Fig. \ref{fig3}, Uber-FuseNet tends to generate over-smoothed results in regions containing rich structures, which implies spatial geometric constraints are not explored well, where the visual features dominate the final estimation in some regions with heavy illuminations and shadows. Instead, our method is more robust and generates the finer depths with crisp objects' boundaries in both near and distant areas.

\textbf{Results on NYUv2 Dataset}. Furthermore, we perform experiments on NYUv2 dataset without any specific modifications. Following existing methods \cite{Ma2018SparsetoDenseDP,Zhang2018DeepDC}, we train and evaluate the performance of our approach with the setting of 200 and 500 sparse LiDAR samples separately. The quantitative results are reported in Table \ref{tab2}, the proposed method achieve the significant gains by integrating extra geometric embedding (denoted as \textbf{w/ DGR}), and more samples bring more gains. However, removing DGR module (denoted as \textbf{w/o DGR}), our method still achieves competitive results, which implies that our framework is effective. 

Furthermore, we select two representative methods, i.e. Sparse2Dense \cite{Ma2018SparsetoDenseDP} and CSPN \cite{Cheng2018DepthEV}, for qualitative comparison. As illuminated in Fig. \ref{fig4}, our method presents superior ability on fine depth estimation.

\subsection{Ablation Study}
The extensive ablation studies are performed on KITTI validation set to analyze the effectiveness of the proposed method. Specifically, we mainly explore two modules, i.e. dynamic graph representation (DGR) and geometry-guided propagation. We first compare each variant of our method and then analyze the generalization ability and stability of model with different network configurations. Note that all variants are trained and evaluated with the same hyperparameters for fair comparison.

\textbf{Components Analysis}. We use the modified ERFNet as our baseline model, which takes RGB-D as input directly. For variants, we sequentially add our modifications into the baseline to verify the effect of each component.

Bi-branch encoding (BBE) is first explored, where the $E_C$ and $E_G$ take the RGB and depth images as input respectively. For DGR module, we consider integrating it into our framework in different manner, e.g., single-scale guidance (SG) instead of multi-scale guidance. So that we only refine final depth in last layer with a fusion block, this is similar with Uber-FuseNet \cite{Chen2019LearningJ2}. In addition, we also explore the effects of local and global feature aggregation operation (reference as FA) and different sampling way, e.g., the farthest point sampling (FPS). We remove FA to verify the performance of model. For FPS, we compare it with random sampling (RS) in our full model configuration. Detailed results are reported in Table \ref{tab3}.

\begin{table}[t]
	\begin{center}
		\begin{tabular}{|l|cccc|} 
			\hline
			$k$-Nearest Neighbors & 3 & 6 & 9 & 12\\
			\hline
			RMSE & $820.12$& $825.37$ & \textbf{813.83} & $823.92$ \\
			\hline
		\end{tabular}
	\end{center}
	\caption{Results of our model with different numbers of $k$ nearest neighbors on KITTI selected validation set.
	}
	\label{tab4}
\end{table}
\begin{figure}[t]
	\begin{center}
		\begin{minipage}[t]{0.49\linewidth}
			\centerline{\includegraphics[width=\textwidth]{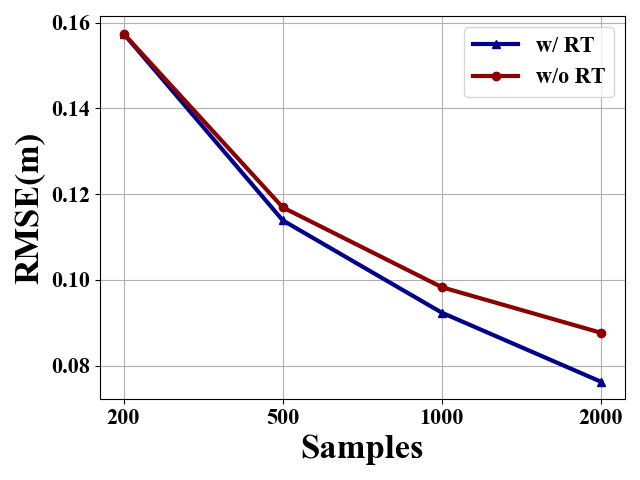}}
			\centerline{(a)}
		\end{minipage}
		\begin{minipage}[t]{0.49\linewidth}
			\centerline{\includegraphics[width=\textwidth]{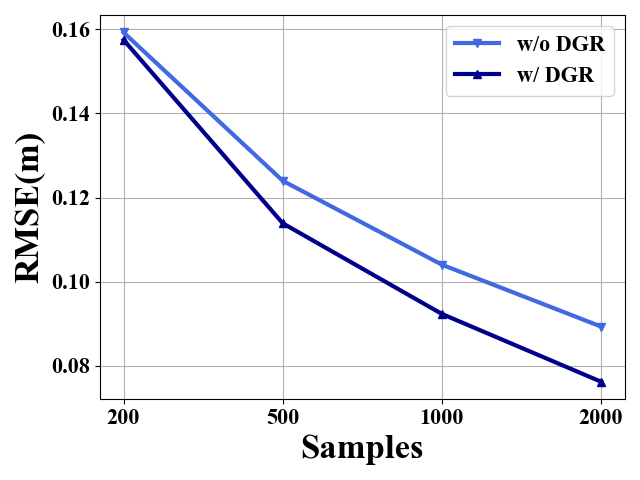}}
			\centerline{(b)}
		\end{minipage}
	\end{center}
	\caption{Generalization analysis on NYUv2 test set. (a) The result of models with 200 samples training and testing with more points, where ``RT" means retraining. (b) The effects of DGR module for performance of model, where ``w/ DRG" denotes the model with DGR module.}
	\label{fig5}
\end{figure}

BBE increases the ability of baseline model, and single-scale guided DGR also improves the performance of the model but with few gains. Similarly, Uber-FuseNet \cite{Chen2019LearningJ2} stacks more fusion modules to improve the performance, it leads to heavy computational cost. Instead, our full model only utilizes single DGR module and achieves significant gains with few training parameters (0.126M only). In addition, removing FA operation decreases the partial ability of the model, which implies it is more robust to aggregate local and global geometric structure information for complex scene perception. Further, we observe that the FPS strategy couldn't provide better performance than random sampling, while reducing the training efficiency with heavily.

\textbf{Receptive filed of DGR}. DGR module mainly relies on $k$-nearest neighbor selection to construct graph network, and enlarges the receptive filed to capture structural priors over the 3D space. Thus, we study the different $k$ setting to explore the effect of receptive filed. Results are shown in Table \ref{tab4}, we select $k=9$ to construct our model.

\begin{figure}[t]
	\begin{center}
		\begin{minipage}[t]{0.24\linewidth}
			\centerline{\includegraphics[width=\textwidth]{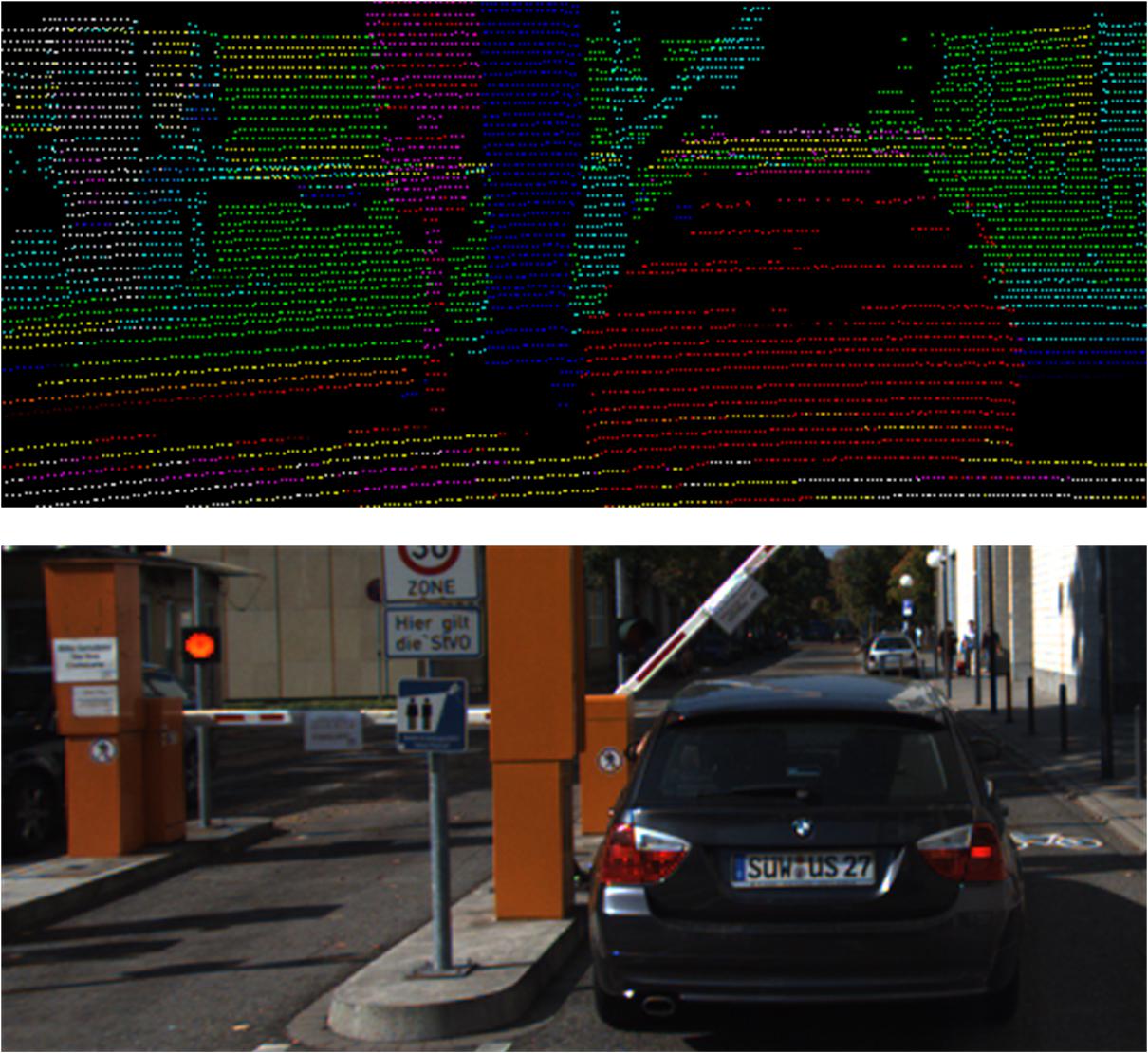}}
			\centerline{Input}
		\end{minipage}
		\begin{minipage}[t]{0.24\linewidth}
			\centerline{\includegraphics[width=\textwidth]{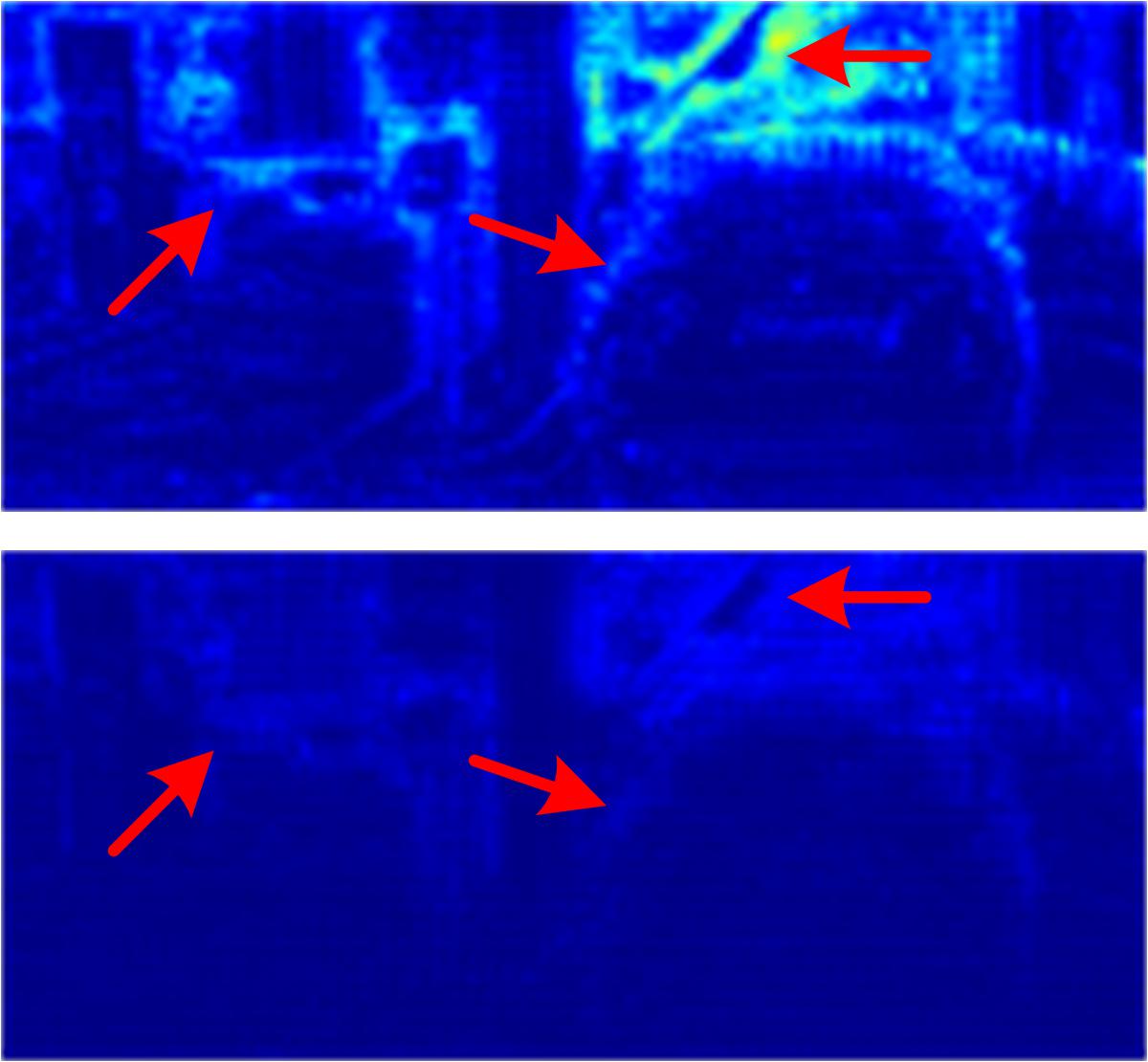}}
			\centerline{1/4-scale}
		\end{minipage}
		\begin{minipage}[t]{0.24\linewidth}
			\centerline{\includegraphics[width=\textwidth]{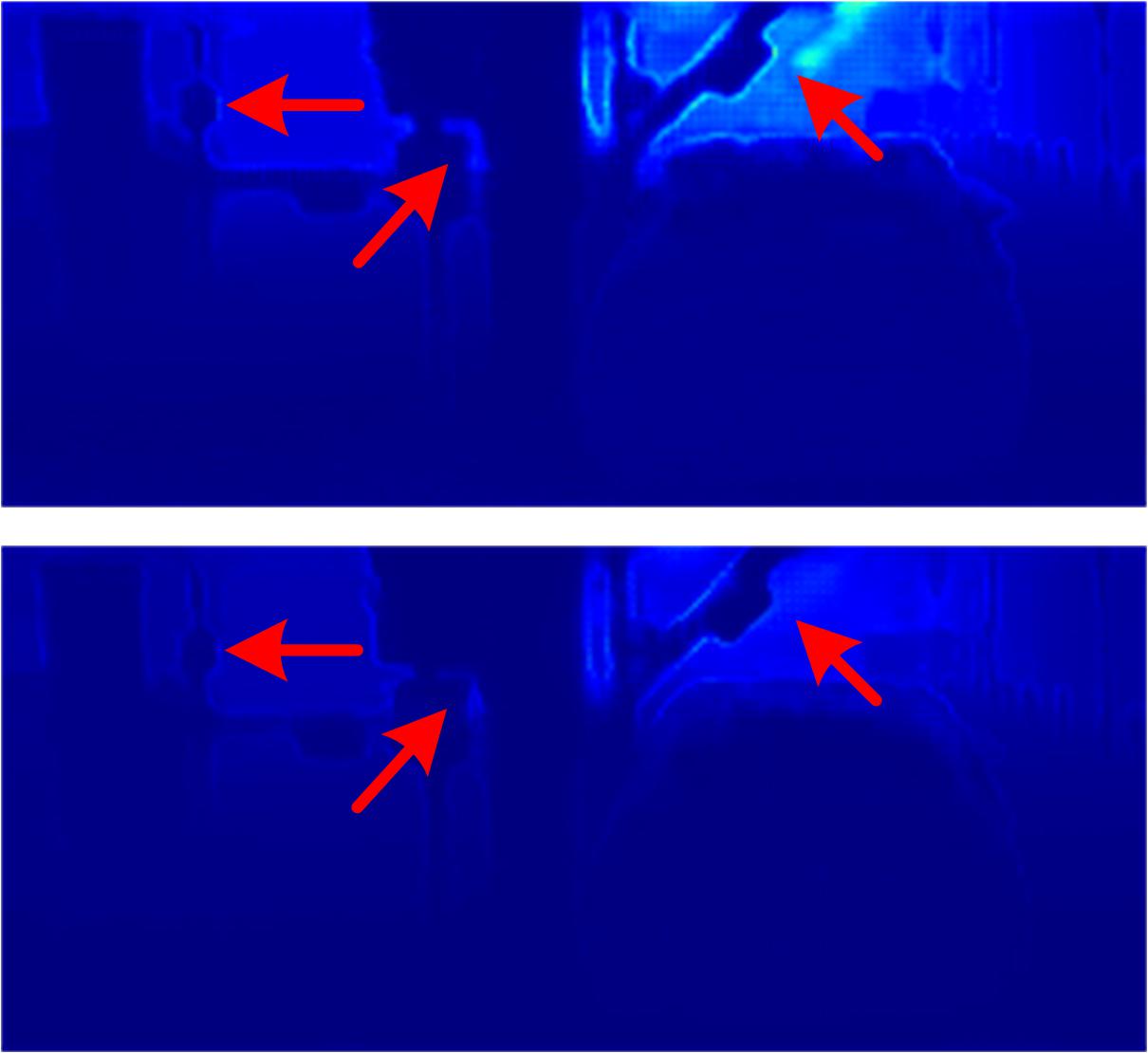}}
			\centerline{1/2-scale}
		\end{minipage}
		\begin{minipage}[t]{0.24\linewidth}
			\centerline{\includegraphics[width=\textwidth]{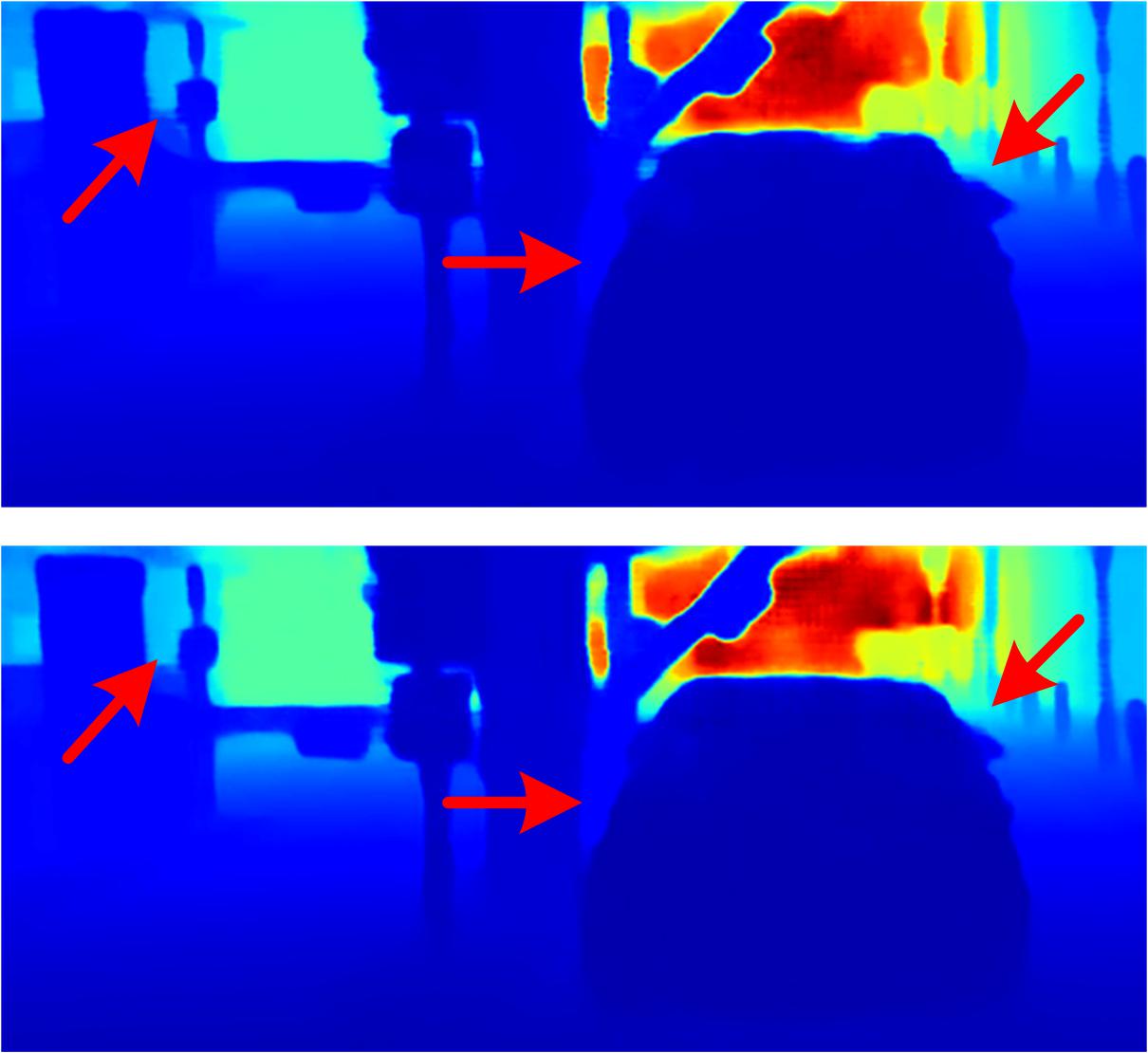}}
			\centerline{Output}
		\end{minipage}
		
	\end{center}
	\caption{Visualized feature maps from upsampling. The first row takes the RGB-Embedding as input, the second row only utilizes the RGB-Depth as input. Zoomed in for better visualization.}
	\label{fig6}
\end{figure}

\begin{figure}[t]
	\begin{center}
		\begin{minipage}[t]{0.156\linewidth}
			\centerline{\includegraphics[width=\textwidth]{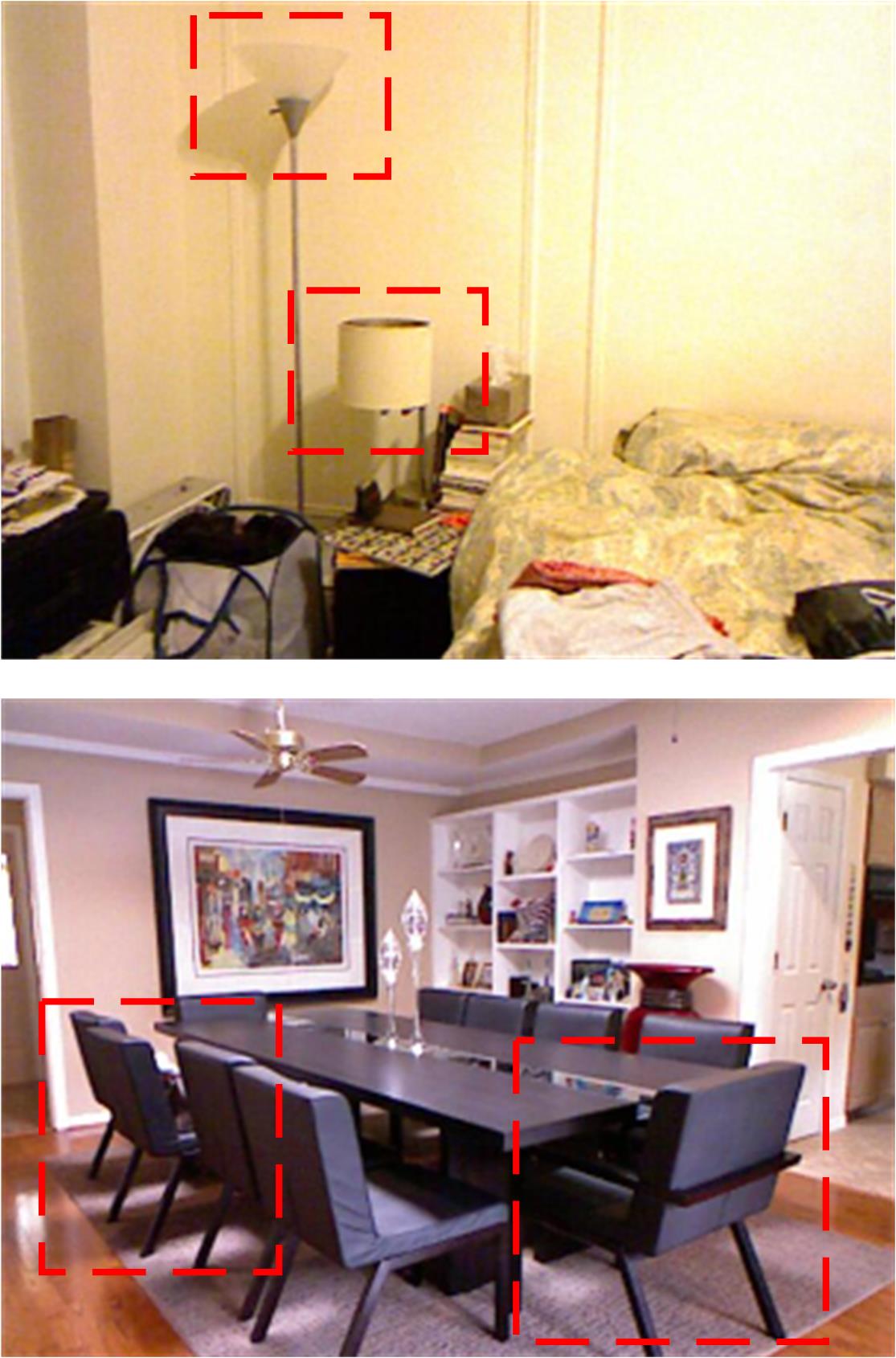}}
			\centerline{RGB}
		\end{minipage}
		\begin{minipage}[t]{0.156\linewidth}
			\centerline{\includegraphics[width=\textwidth]{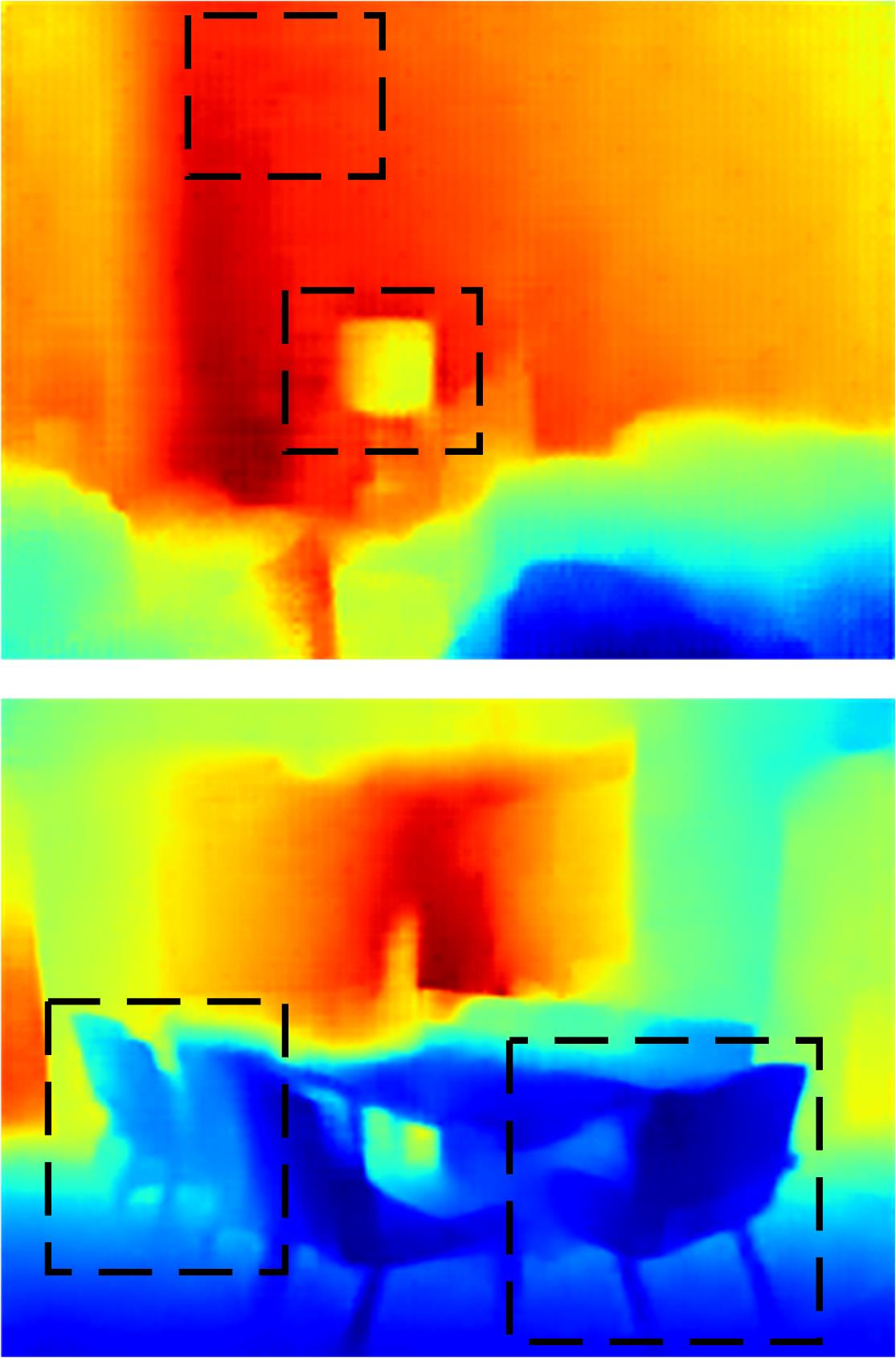}}
			\centerline{w/o 200}
		\end{minipage}
		\begin{minipage}[t]{0.156\linewidth}
			\centerline{\includegraphics[width=\textwidth]{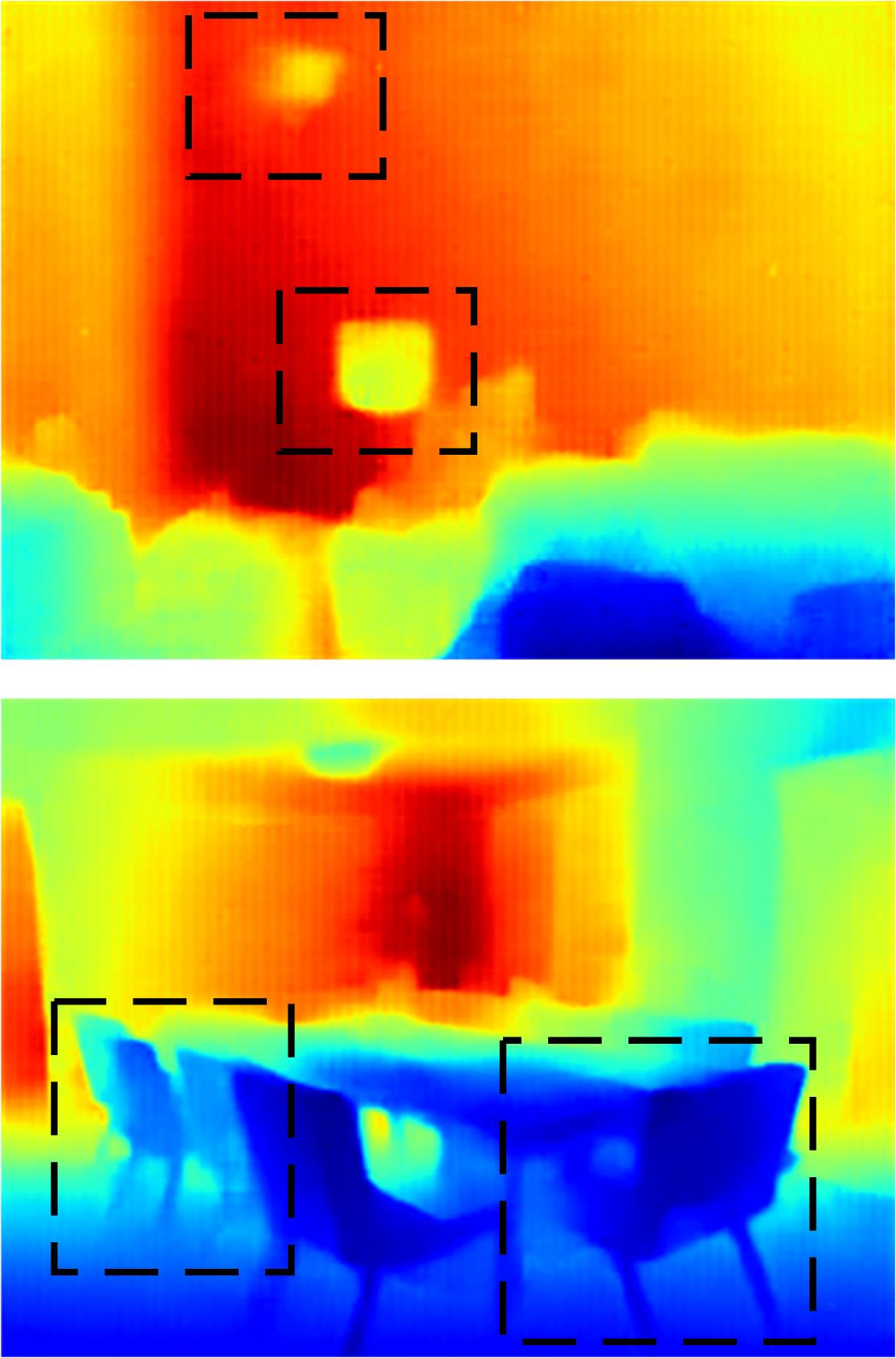}}
			\centerline{w/ 200}
		\end{minipage}
		\begin{minipage}[t]{0.156\linewidth}
			\centerline{\includegraphics[width=\textwidth]{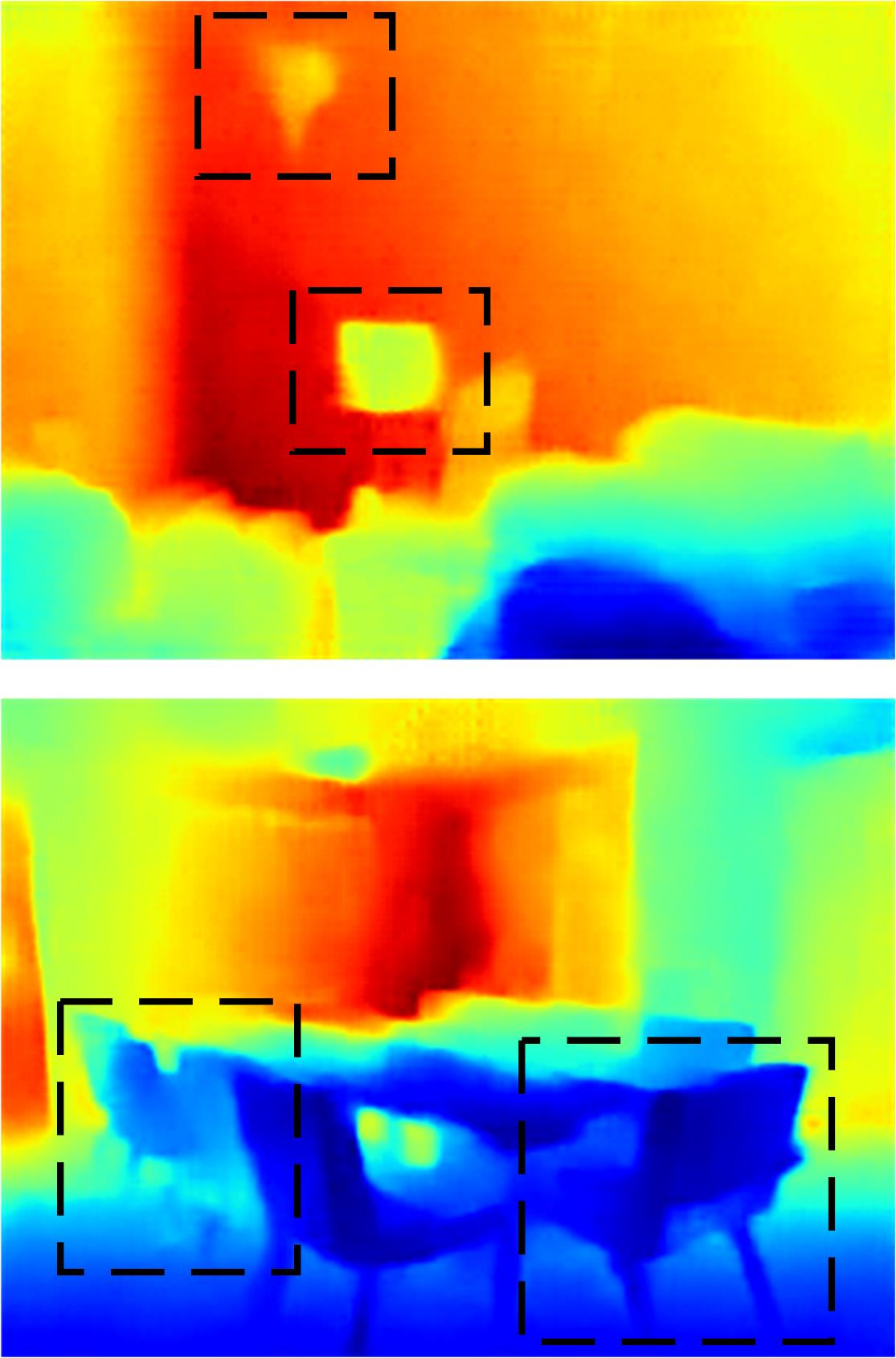}}
			\centerline{w/o 500}
		\end{minipage}
		\begin{minipage}[t]{0.156\linewidth}
			\centerline{\includegraphics[width=\textwidth]{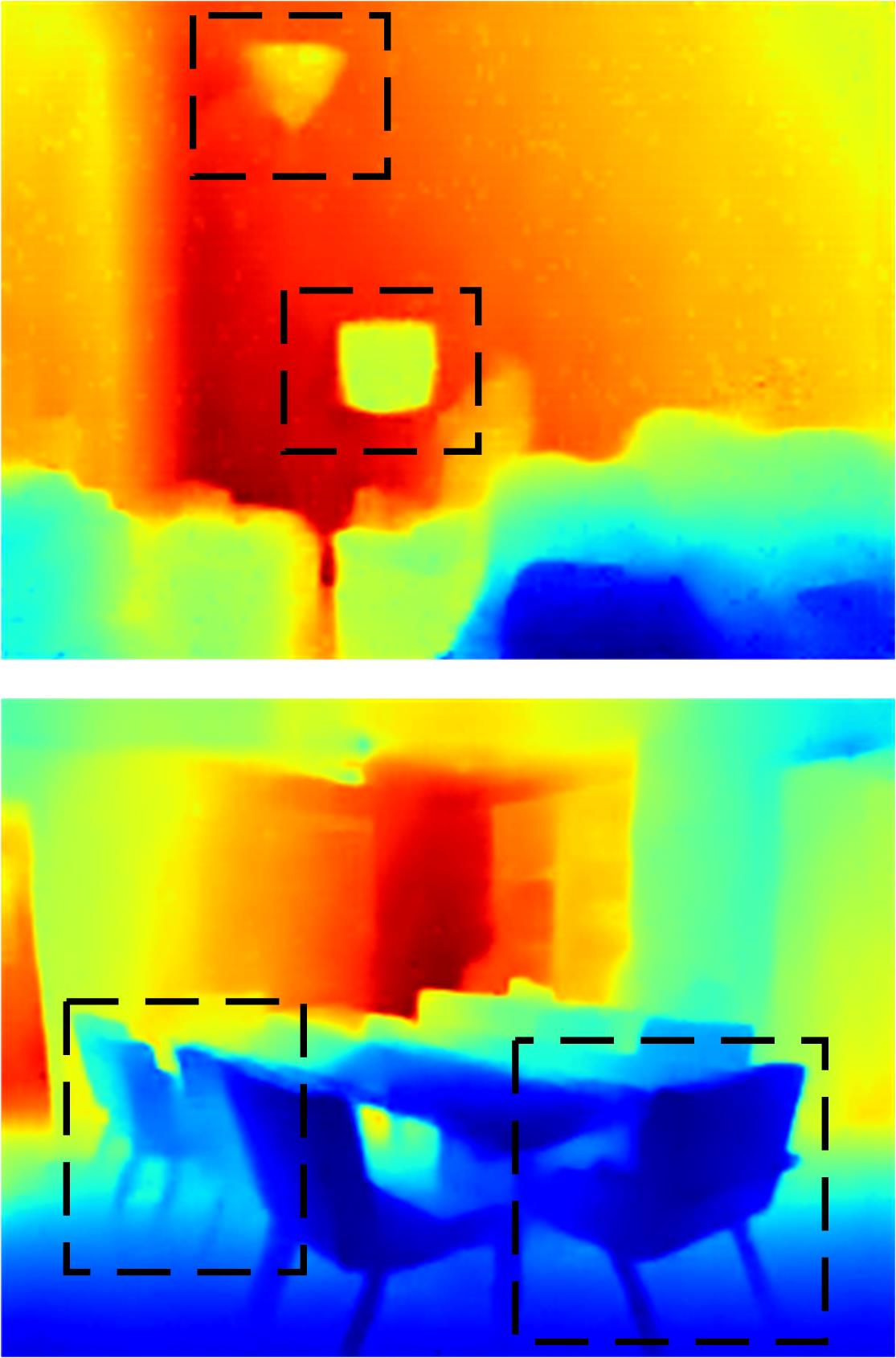}}
			\centerline{w/ 500}
		\end{minipage}
		\begin{minipage}[t]{0.156\linewidth}
			\centerline{\includegraphics[width=\textwidth]{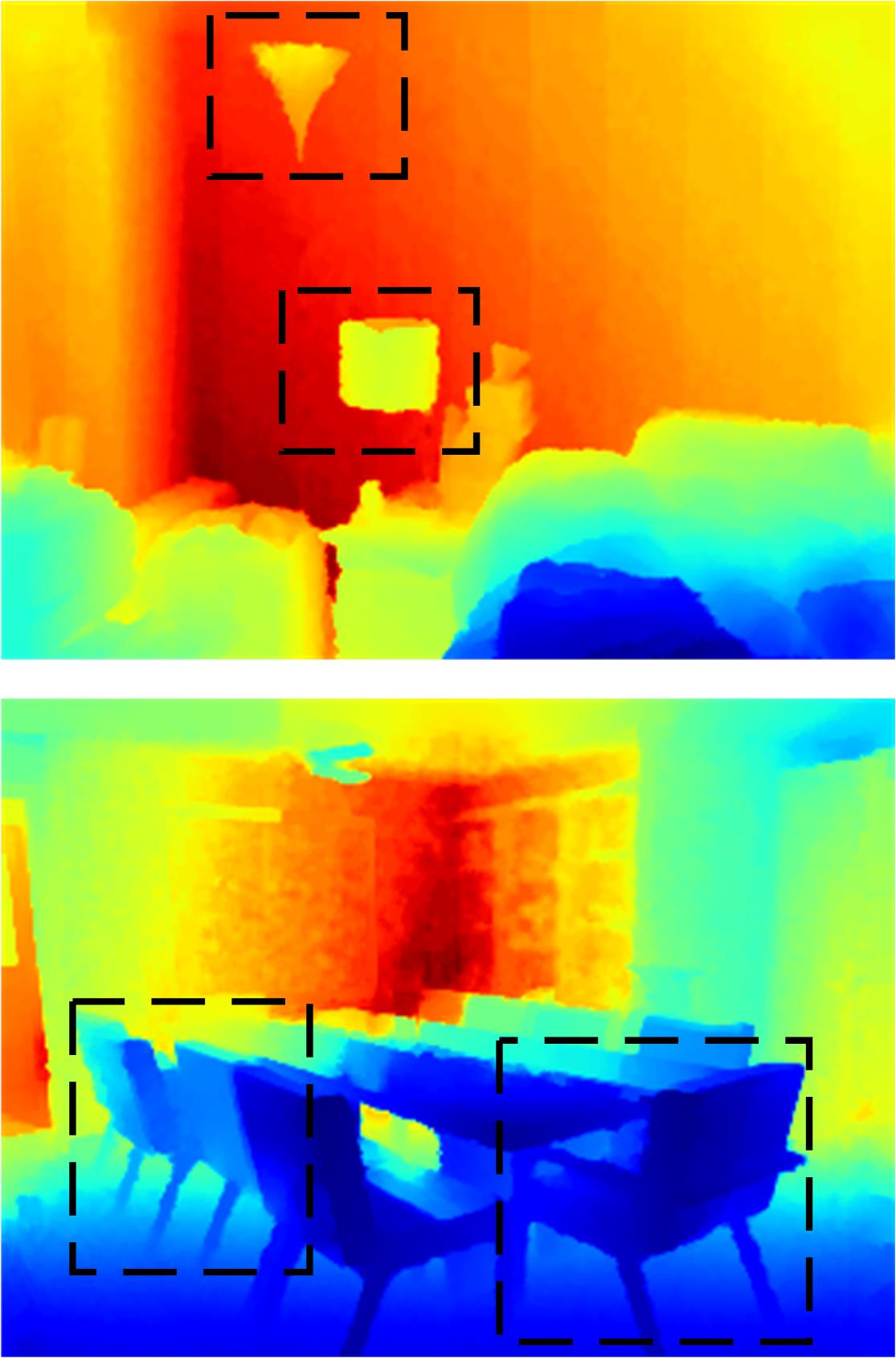}}
			\centerline{GT}
		\end{minipage}
	\end{center}
	\caption{The effect of DGR for visualized results. Implicit geometric constraints provide finer estimation, where ``w/o" means removing DGR module, ``200" denotes using 200 sparse LiDAR samples.}
	\label{fig7}
\end{figure}
\textbf{Generalization capability}. Our DGR module is trained with fixed-number points in KITTI dataset, which losses partial information of the original point cloud. Therefore, the goal of DGR is to capture generalized representation from uncertain inputs so that geometric clues could be utilized sufficiently from limited points. To this end, we evaluate the model with different numbers of LiDARs on NYUv2 benchmark, i.e. training with 200 samples and testing with more samples, e.g., 500, 1000, and 2000. The results are shown in Fig. \ref{fig5}(a). As LiDARs increasing, our approach achieves significant gains without retrained, which presents consistent improvements with models based on retrained with fixed points. This implies that DGR could generalize on uncertain point clouds well. Further, we compare the performance of models removing DGR module. As shown in Fig. \ref{fig5}(b), DGR brings significant gains, and more points more gains, which implies the effectiveness of our method. 

\textbf{Implicit Geometric Constraints.} We also explore the effect of DGR in guided propagation module. Visualized upsampled feature maps from KITTI are illuminated in Fig. \ref{fig6}. The outputs injected geometric embedding present more structural responses on objects' edges and scene layouts. It supports recent researches on structural scene depth estimation \cite{Hu2019VisualizationOC,Chen2021S2RDepthNetLA}. Further, more visualized results from NYUv2 are shown in Fig. \ref{fig7}.

\section{CONCLUSIONS}

In this paper, we have presented a simple yet efficient depth completion method, which integrates 3D geometric representation into the 2D learning architecture and achieves a better trade-off between the performance and efficiency. While many concrete implementations of the general idea, including utilize powerful graph networks and complicated fusion frameworks, are possible, we show that a simple design already achieves competing results. Experiments on the outdoor and indoor scenes demonstrate our approach could predict crisp depths while retaining structure and scale consistency. 


\addtolength{\textheight}{-1cm}  




{\small
	\bibliographystyle{IEEEtran}
	\bibliography{main}

\begin{thebibliography}{10}
\providecommand{\url}[1]{#1}
\csname url@rmstyle\endcsname
\providecommand{\newblock}{\relax}
\providecommand{\bibinfo}[2]{#2}
\providecommand\BIBentrySTDinterwordspacing{\spaceskip=0pt\relax}
\providecommand\BIBentryALTinterwordstretchfactor{4}
\providecommand\BIBentryALTinterwordspacing{\spaceskip=\fontdimen2\font plus
\BIBentryALTinterwordstretchfactor\fontdimen3\font minus
  \fontdimen4\font\relax}
\providecommand\BIBforeignlanguage[2]{{%
\expandafter\ifx\csname l@#1\endcsname\relax
\typeout{** WARNING: IEEEtran.bst: No hyphenation pattern has been}%
\typeout{** loaded for the language `#1'. Using the pattern for}%
\typeout{** the default language instead.}%
\else
\language=\csname l@#1\endcsname
\fi
#2}}

\bibitem{Szeliski2011ComputerV}
R.~Szeliski, ``Computer vision - algorithms and applications,'' in \emph{Texts
  in Computer Science}, 2011.

\bibitem{Uhrig2017SparsityIC}
J.~Uhrig, N.~Schneider, L.~Schneider, U.~Franke, T.~Brox, and A.~Geiger,
  ``Sparsity invariant cnns,'' \emph{2017 International Conference on 3D Vision
  (3DV)}, pp. 11--20, 2017.

\bibitem{Ma2018SparsetoDenseDP}
F.~Ma and S.~Karaman, ``Sparse-to-dense: Depth prediction from sparse depth
  samples and a single image,'' \emph{2018 IEEE International Conference on
  Robotics and Automation (ICRA)}, pp. 1--8, 2018.

\bibitem{Gansbeke2019SparseAN}
W.~V. Gansbeke, D.~Neven, B.~D. Brabandere, and L.~Gool, ``Sparse and noisy
  lidar completion with rgb guidance and uncertainty,'' \emph{2019 16th
  International Conference on Machine Vision Applications (MVA)}, pp. 1--6,
  2019.

\bibitem{Eldesokey2020ConfidencePT}
A.~Eldesokey, M.~Felsberg, and F.~Khan, ``Confidence propagation through cnns
  for guided sparse depth regression,'' \emph{IEEE Transactions on Pattern
  Analysis and Machine Intelligence}, vol.~42, pp. 2423--2436, 2020.

\bibitem{Zhang2018DeepDC}
Y.~Zhang and T.~Funkhouser, ``Deep depth completion of a single rgb-d image,''
  \emph{2018 IEEE/CVF Conference on Computer Vision and Pattern Recognition},
  pp. 175--185, 2018.

\bibitem{Lee2019DepthCW}
B.~uk~Lee, H.-G. Jeon, S.~Im, and I.~S. Kweon, ``Depth completion with deep
  geometry and context guidance,'' \emph{2019 International Conference on
  Robotics and Automation (ICRA)}, pp. 3281--3287, 2019.

\bibitem{Qiu2019DeepLiDARDS}
J.~Qiu, Z.~Cui, Y.~Zhang, X.~Zhang, S.~Liu, B.~Zeng, and M.~Pollefeys,
  ``Deeplidar: Deep surface normal guided depth prediction for outdoor scene
  from sparse lidar data and single color image,'' \emph{2019 IEEE/CVF
  Conference on Computer Vision and Pattern Recognition (CVPR)}, pp.
  3308--3317, 2019.

\bibitem{Xu2019DepthCF}
Y.~Xu, X.~Zhu, J.~Shi, G.~Zhang, H.~Bao, and H.~Li, ``Depth completion from
  sparse lidar data with depth-normal constraints,'' \emph{2019 IEEE/CVF
  International Conference on Computer Vision (ICCV)}, pp. 2811--2820, 2019.

\bibitem{Chen2019LearningJ2}
Y.~Chen, B.~Yang, M.~Liang, and R.~Urtasun, ``Learning joint 2d-3d
  representations for depth completion,'' \emph{2019 IEEE/CVF International
  Conference on Computer Vision (ICCV)}, pp. 10\,022--10\,031, 2019.

\bibitem{Xiang20203dDepthNetPC}
R.~Xiang, F.~Zheng, H.~Su, and Z.~Zhang, ``3ddepthnet: Point cloud guided depth
  completion network for sparse depth and single color image,'' \emph{ArXiv},
  vol. abs/2003.09175, 2020.

\bibitem{Xiong2020SparsetoDenseDC}
X.~Xiong, H.~Xiong, K.~Xian, C.~Zhao, Z.~Cao, and X.~Li, ``Sparse-to-dense
  depth completion revisited: Sampling strategy and graph construction,'' in
  \emph{ECCV}, 2020.

\bibitem{Hawe2011DenseDM}
S.~Hawe, M.~Kleinsteuber, and K.~Diepold, ``Dense disparity maps from sparse
  disparity measurements,'' \emph{2011 International Conference on Computer
  Vision}, pp. 2126--2133, 2011.

\bibitem{Liu2015DepthRF}
L.-K. Liu, S.~Chan, and T.~Q. Nguyen, ``Depth reconstruction from sparse
  samples: Representation, algorithm, and sampling,'' \emph{IEEE Transactions
  on Image Processing}, vol.~24, pp. 1983--1996, 2015.

\bibitem{Huang2020HMSNetHM}
Z.~Huang, J.~Fan, S.~Cheng, S.~Yi, X.~Wang, and H.~Li, ``Hms-net: Hierarchical
  multi-scale sparsity-invariant network for sparse depth completion,''
  \emph{IEEE Transactions on Image Processing}, vol.~29, pp. 3429--3441, 2020.

\bibitem{Cheng2018DepthEV}
X.~Cheng, P.~Wang, and R.~Yang, ``Depth estimation via affinity learned with
  convolutional spatial propagation network,'' in \emph{ECCV}, 2018.

\bibitem{Liu2017LearningAV}
S.~Liu, S.~D. Mello, J.~Gu, G.~Zhong, M.-H. Yang, and J.~Kautz, ``Learning
  affinity via spatial propagation networks,'' in \emph{NIPS}, 2017.

\bibitem{Park2020NonLocalSP}
J.~Park, K.~Joo, Z.~Hu, C.-K. Liu, and I.~Kweon, ``Non-local spatial
  propagation network for depth completion,'' in \emph{ECCV}, 2020.

\bibitem{Cheng2020CSPNLC}
X.~Cheng, P.~Wang, C.~Guan, and R.~Yang, ``Cspn++: Learning context and
  resource aware convolutional spatial propagation networks for depth
  completion,'' \emph{AAAI}, 2020.

\bibitem{Hu2021PENetTP}
M.~Hu, S.~Wang, B.~Li, S.~Ning, L.~Fan, and X.~Gong, ``Penet: Towards precise
  and efficient image guided depth completion,'' \emph{2021 IEEE International
  Conference on Robotics and Automation (ICRA)}, 2021.

\bibitem{Shivakumar2019DFuseNetDF}
S.~S. Shivakumar, T.~Nguyen, S.~W. Chen, and C.~J. Taylor, ``Dfusenet: Deep
  fusion of rgb and sparse depth information for image guided dense depth
  completion,'' \emph{2019 IEEE Intelligent Transportation Systems Conference
  (ITSC)}, pp. 13--20, 2019.

\bibitem{Tang2021LearningGC}
J.~Tang, F.-P. Tian, W.~Feng, J.~Li, and P.~Tan, ``Learning guided
  convolutional network for depth completion,'' \emph{IEEE Transactions on
  Image Processing}, vol.~30, pp. 1116--1129, 2021.

\bibitem{Imran2021DepthCW}
S.~Imran, X.~Liu, and D.~Morris, ``Depth completion with twin surface
  extrapolation at occlusion boundaries,'' \emph{CVPR}, 2021.

\bibitem{Lee2021DepthCU}
B.-U. Lee, K.~Lee, and I.~Kweon, ``Depth completion using plane-residual
  representation,'' in \emph{CVPR}, 2021.

\bibitem{Ma2019SelfSupervisedSS}
F.~Ma, G.~Cavalheiro, and S.~Karaman, ``Self-supervised sparse-to-dense:
  Self-supervised depth completion from lidar and monocular camera,''
  \emph{2019 International Conference on Robotics and Automation (ICRA)}, pp.
  3288--3295, 2019.

\bibitem{Wong2020UnsupervisedDC}
A.~Wong, X.~Fei, S.~Tsuei, and S.~Soatto, ``Unsupervised depth completion from
  visual inertial odometry,'' \emph{IEEE Robotics and Automation Letters},
  vol.~5, pp. 1899--1906, 2020.

\bibitem{Wang2018DeepPC}
S.~Wang, S.~Suo, W.-C. Ma, A.~Pokrovsky, and R.~Urtasun, ``Deep parametric
  continuous convolutional neural networks,'' \emph{2018 IEEE/CVF Conference on
  Computer Vision and Pattern Recognition}, pp. 2589--2597, 2018.

\bibitem{Wu2019PointConvDC}
W.~Wu, Z.~Qi, and F.~Li, ``Pointconv: Deep convolutional networks on 3d point
  clouds,'' \emph{2019 IEEE/CVF Conference on Computer Vision and Pattern
  Recognition (CVPR)}, pp. 9613--9622, 2019.

\bibitem{Su2015MultiviewCN}
H.~Su, S.~Maji, E.~Kalogerakis, and E.~Learned-Miller, ``Multi-view
  convolutional neural networks for 3d shape recognition,'' \emph{2015 IEEE
  International Conference on Computer Vision (ICCV)}, pp. 945--953, 2015.

\bibitem{Klokov2017EscapeFC}
R.~Klokov and V.~Lempitsky, ``Escape from cells: Deep kd-networks for the
  recognition of 3d point cloud models,'' \emph{2017 IEEE International
  Conference on Computer Vision (ICCV)}, pp. 863--872, 2017.

\bibitem{Qi2017PointNetDL}
C.~Qi, H.~Su, K.~Mo, and L.~Guibas, ``Pointnet: Deep learning on point sets for
  3d classification and segmentation,'' \emph{2017 IEEE Conference on Computer
  Vision and Pattern Recognition (CVPR)}, pp. 77--85, 2017.

\bibitem{Qi2017PointNetDH}
C.~Qi, L.~Yi, H.~Su, and L.~Guibas, ``Pointnet++: Deep hierarchical feature
  learning on point sets in a metric space,'' in \emph{NIPS}, 2017.

\bibitem{Te2018RGCNNRG}
G.~Te, W.~Hu, Z.~Guo, and A.~Zheng, ``Rgcnn: Regularized graph cnn for point
  cloud segmentation,'' \emph{Proceedings of the 26th ACM international
  conference on Multimedia}, 2018.

\bibitem{Wang2019DynamicGC}
Y.~Wang, Y.~Sun, Z.~Liu, S.~E. Sarma, M.~Bronstein, and J.~Solomon, ``Dynamic
  graph cnn for learning on point clouds,'' \emph{ACM Transactions on Graphics
  (TOG)}, vol.~38, pp. 1 -- 12, 2019.

\bibitem{Zhang2019LinkedDG}
K.~Zhang, M.~Hao, J.~Wang, C.~W. de~Silva, and C.~Fu, ``Linked dynamic graph
  cnn: Learning on point cloud via linking hierarchical features,''
  \emph{ArXiv}, vol. abs/1904.10014, 2019.

\bibitem{Zhao2021AdaptiveCM}
S.~Zhao, M.~Gong, H.~Fu, and D.~Tao, ``Adaptive context-aware multi-modal
  network for depth completion,'' \emph{IEEE Transactions on Image Processing},
  vol.~30, pp. 5264--5276, 2021.

\bibitem{Zhao2020MonocularDE}
C.~Zhao, Q.~Sun, C.~Zhang, Y.~Tang, and F.~Qian, ``Monocular depth estimation
  based on deep learning: An overview,'' \emph{Science China Technological
  Sciences}, pp. 1--16, 2020.

\bibitem{Jaritz2018SparseAD}
M.~Jaritz, R.~de~Charette, {\'E}.~Wirbel, X.~Perrotton, and F.~Nashashibi,
  ``Sparse and dense data with cnns: Depth completion and semantic
  segmentation,'' \emph{2018 International Conference on 3D Vision (3DV)}, pp.
  52--60, 2018.

\bibitem{romera2018erfnet}
E.~{Romera}, J.~M. {Alvarez}, L.~M. {Bergasa}, and R.~{Arroyo}, ``Erfnet:
  Efficient residual factorized convnet for real-time semantic segmentation,''
  \emph{IEEE Transactions on Intelligent Transportation Systems}, vol.~19,
  no.~1, pp. 263--272, 2018.

\bibitem{Sandler2018MobileNetV2IR}
M.~Sandler, A.~G. Howard, M.~Zhu, A.~Zhmoginov, and L.-C. Chen, ``Mobilenetv2:
  Inverted residuals and linear bottlenecks,'' \emph{2018 IEEE/CVF Conference
  on Computer Vision and Pattern Recognition}, pp. 4510--4520, 2018.

\bibitem{Silberman2012IndoorSA}
N.~Silberman, D.~Hoiem, P.~Kohli, and R.~Fergus, ``Indoor segmentation and
  support inference from rgbd images,'' in \emph{ECCV}, 2012.

\bibitem{Kingma2015AdamAM}
D.~P. Kingma and J.~Ba, ``Adam: A method for stochastic optimization,''
  \emph{CoRR}, vol. abs/1412.6980, 2015.

\bibitem{Ferstl2013ImageGD}
D.~Ferstl, C.~Reinbacher, R.~Ranftl, M.~R{\"u}ther, and H.~Bischof, ``Image
  guided depth upsampling using anisotropic total generalized variation,''
  \emph{2013 IEEE International Conference on Computer Vision}, pp. 993--1000,
  2013.

\bibitem{Hu2019VisualizationOC}
J.~Hu, Y.~Zhang, and T.~Okatani, ``Visualization of convolutional neural
  networks for monocular depth estimation,'' \emph{2019 IEEE/CVF International
  Conference on Computer Vision (ICCV)}, pp. 3868--3877, 2019.

\bibitem{Chen2021S2RDepthNetLA}
X.~Chen, Y.~Wang, X.~Chen, and W.~Zeng, ``S2r-depthnet: Learning a
  generalizable depth-specific structural representation,'' in \emph{CVPR},
  2021.

\end{thebibliography}
}
\end{document}